\documentclass[journal,twoside,web]{ieeecolor}
\usepackage{generic}
\usepackage{cite}
\usepackage{amsmath,amssymb,amsfonts}
\usepackage{algorithmic}
\usepackage{graphicx}
\usepackage{algorithm,algorithmic}
\usepackage{hyperref}
\hypersetup{hidelinks=true}
\usepackage{textcomp}
\usepackage{tabularray}
\usepackage{placeins}
\usepackage{generic}
\usepackage{latexsym}
\usepackage{graphicx}
\usepackage{amsfonts,amssymb,amsmath}
\usepackage{cite}
\usepackage{textcomp}
\usepackage{booktabs}
\usepackage{multirow}
\usepackage{float}
\usepackage[switch]{lineno}
\usepackage{arydshln}
\usepackage{cite}
\usepackage{url}
\usepackage{amsmath,amssymb,amsfonts}
\usepackage{algorithmic}
\usepackage{mathtools} 
\usepackage{multirow}
\usepackage{graphicx}
\usepackage{scrextend}
\usepackage{xr-hyper}

\makeatletter
\newcommand*{\addFileDependency}[1]{
  \typeout{(#1)}
  \@addtofilelist{#1}
  \IfFileExists{#1}{}{\typeout{No file #1.}}
}

\newcommand{\hly}[1]{\textcolor{black}{#1}}
\newcommand{\hlb}[1]{\textcolor{black}{#1}}
\makeatother

\newcommand*{\myexternaldocument}[1]{
    \externaldocument{#1}
    \addFileDependency{#1.tex}
    \addFileDependency{#1.aux}
}

\myexternaldocument{supply_clean}
\def\BibTeX{{\rm B\kern-.05em{\sc i\kern-.025em b}\kern-.08em
    T\kern-.1667em\lower.7ex\hbox{E}\kern-.125emX}}
\begin{document}
\title{KidMesh: Computational Mesh Reconstruction for Pediatric Congenital Hydronephrosis Using Deep Neural Networks}
\author{Haoran Sun, Zhanpeng Zhu, Anguo Zhang, Bo Liu, Zhaohua Lin, Liqin Huang, Mingjing Yang, Lei Liu, Shan Lin, and Wangbin Ding
\thanks{This work was supported by the National Natural Science Foundation of China (62401148, 62271149), Fuzhou Science and Technology Planning Project (2023-P-001), Fujian Science and Technology Funds (2025J01072, 2023Y9308, 2024J01353, 2022Y0056, 2022Y4014, 2020Y9091).
(Corresponding authors: Mingjing Yang, Lei Liu, Shan Lin, and Wangbin Ding.)}
\thanks{Haoran Sun and Lei Liu are with the School of Basic Medical Sciences, Intelligent Medicine Institute, Fudan University, Shanghai 200032, China (e-mail: manglu3935@126.com; liulei@fudan.edu.cn).}
\thanks{Wangbin Ding and Zhaohua Lin are with the School of Medical Imaging, Fujian Medical University, Fuzhou 350100, China and also with the Virtual Teaching and Research Group for Intelligent Precision Radiotherapy, Fujian Medical University, Fuzhou 350100, China (e-mail: dingwangbin@126.com; zhlin2328@gmail.com).}
\thanks{Zhanpeng Zhu, Liqin Huang and Mingjing Yang are with the College of Physics and Information Engineering, Fuzhou University, Fuzhou 350108, China (e-mail: 231127178@fzu.edu.cn; hlq@fzu.edu.cn; yangmj5@fzu.edu.cn).}
\thanks{Shan Lin is with the department of pediatrics,  Fuzhou University Affiliated Provincial Hospital, No.134 East street, Fuzhou 350108, Fujian, China and also with  Shengli Clinical Medical College, Fujian Medical University, Fujian Provincial Hospital, No.134 East street, Fuzhou 350108, Fujian, China (e-mail: Linshan@fzu.edu.cn).
}
\thanks{Anguo Zhang is with the Interdisciplinary Institute of Medical Engineering, Fuzhou University, Fuzhou 350108, China (e-mail: anguo.zhang@hotmail.com).}
\thanks{Bo Liu is with the Department of Thoracic Surgery, The First Affiliated Hospital, Fujian Medical University, Fuzhou 350005, China (e-mail: 328221720@qq.com).}}

\maketitle

\begin{abstract}
Pediatric congenital hydronephrosis (CH) is a common  urinary tract disorder, primarily caused by obstruction at the renal pelvis-ureter junction. Magnetic resonance urography (MRU) can visualize hydronephrosis, including renal pelvis and calyces, by utilizing the natural contrast provided by water.  Existing voxel-based segmentation approaches can extract CH regions from MRU, facilitating disease diagnosis and prognosis. However, these segmentation methods predominantly focus on morphological features, such as size, shape, and structure. To enable functional assessments, such as urodynamic simulations, external complex post-processing steps are required to convert these results into mesh-level representations. To address this limitation, we propose an end-to-end method based on deep neural networks, namely KidMesh, which could automatically reconstruct CH  meshes directly from MRU. Generally, KidMesh extracts feature maps from MRU images and converts them into feature vertices through grid sampling. It then deforms a template mesh according to these feature vertices to generate the specific CH meshes of MRU images. Meanwhile, we develop a novel schema to train KidMesh without relying on accurate mesh-level annotations, which are difficult to obtain due to the sparsely sampled MRU slices. Experimental results show that KidMesh could reconstruct CH meshes in an average of 0.4 seconds, and achieve comparable performance to conventional methods without requiring post-processing. The reconstructed meshes exhibited no self-intersections, with only 3.7\% and 0.2\% of the vertices having error distances exceeding 3.2mm and 6.4mm, respectively. After rasterization, these meshes achieved a Dice score of 0.86 against manually delineated CH masks.  Furthermore, these meshes could be used in renal urine flow simulations, providing valuable urodynamic information for clinical practice.
\end{abstract}


\section{Introduction}
\label{sec:introduction}
Pediatric congenital hydronephrosis (CH) is often caused by anatomical abnormalities, including ureteral stenosis, atresia, and vesicoureteral reflux \cite{kohno2020pediatric}. The most direct consequence of CH is the swelling of \hlb{the} renal pelvis and calyces, resulting from the accumulation of urine in these areas. It can lead to progressive renal function impairment or hinder the optimal development of kidney function in children \cite{onen2020grading}. Medical imaging, such as ultrasound, renal radionuclide imaging, and magnetic resonance urography (MRU) are common techniques for CH diagnosis and prognosis \cite{Garcia-Valtuille2006,Leyendecker2009}. For instance, ultrasound imaging could visualize the location and degree of hydronephrosis, assisting clinicians in assessing its severity. While MRU offers more detailed anatomical information, enriching the understanding of associated abnormalities. Nevertheless, medical imaging techniques primarily provide structural information, and are limited in assessing urodynamic information for CH.

The development of computational fluid \hlb{dynamics} (CFD) simulation has emerged as a typical solution to the aforementioned limitation. By constructing patient-specific 3D meshes, CFD can infer urodynamic insights for CH. Urologists, specializing in adult care, have begun utilizing the computational models of  renal pelvis and calyces to investigate urodynamic changes caused by obstructive urolithiasis \cite{Qin2021}. These studies predominantly rely on morphological information from previous literature. Most of them are based on images derived from polyester resin corrosion casting of the renal pelvis and calyces from 73 adult cadavers \cite{Morais2024}. However, adult kidney models differ significantly from those of children, and the existing models may not accurately capture the distinct anatomical features of individual pediatric kidneys.

\begin{figure*}[htbp]
    \centering
    \includegraphics[width=0.9\textwidth]{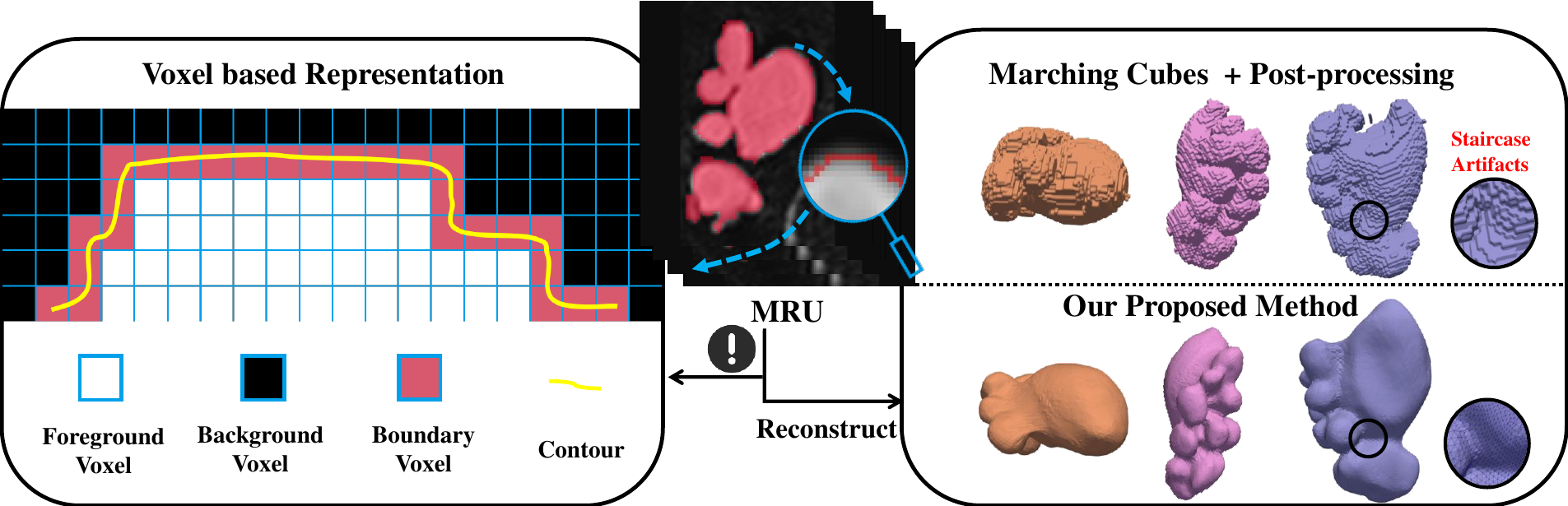}
    \caption{Illustration of \hlb{voxel-based} magnetic resonance urography (MRU) images and mesh reconstruction methods. The left image shows a patch from MRU images, with the yellow contour indicating hydronephrosis boundaries. MRU techniques visualize this region using voxels, often resulting in staircase artifacts (black circles) after reconstruction. The right top panel displays meshes generated using Marching Cubes with post-processing, while the right bottom panel shows our proposed reconstruction meshes based on deep learning.}
    \label{fig1}
\end{figure*}

At present, voxel-based image segmentation is the most common way to reconstruct human organs from medical images for patient-specific CFD simulations \cite{decroocq2023modeling,ferdian2023cerebrovascular,pak2023patient}. Traditional region-growing image segmentation methods often require the manual selection of seed points to initiate segmentation processes. Identifying the appropriate seed points in images can be tedious \cite{Adams1994}. Traditional segmentation methods based on threshold and edge detection can yield precise results. However, they are highly sensitive to image noise, which results in poor robustness \cite{Ju2013,Manjula2017}. With the development of deep learning methods, convolutional neural networks (CNNs) have increasingly been applied to medical image segmentation tasks \cite{Siddique2021}. Various CNN-based network architectures, including multi-view 2D CNN, patch-based 3D CNN, and full-size 3D CNN, have gained prominence in medical segmentation \cite{Ronneberger2015,Zhou2018,Huo2021}. In contrast to traditional methods,  CNN-based approaches can rapidly and automatically learn latent patterns from large datasets, achieving precise segmentation.

Nevertheless, voxel-based segmentation methods require additional isosurface extraction algorithms (e.g., Marching Cubes \cite{Lorensen1998a}) and a series of post-processing techniques \cite{bacciaglia2021surface,charton2021mesh} to generate meshes for CFD simulation. Note that the quality of reconstructed meshes is significantly affected by the resolution of original medical images, as shown in the right panel Figure \ref{fig1}. Due to limitations in medical imaging protocols, such as sparse sampling along the Z-axis, reconstructed meshes often exhibit staircase artifacts \cite{Moench2010}. Even though smoothing filters and topology correction algorithms can refine mesh quality, these processes are not only time-consuming but also do not ensure anatomically accurate meshes \cite{yotter2011topological,wyburd2024anatomically}. 

Recent advancements in deep learning have enabled researchers to explore more effective mesh reconstruction methods. In particular, deep learning approaches have emerged in the realm of cortical and white matter reconstruction. One notable method employed the signed distance function as an implicit representation of the mesh, enabling mesh generation at various desired resolutions \cite{Park2019a}. While this approach mitigates the challenge of original image resolution, it still necessitates essential topology corrections. To address this, pioneering studies have integrated graph convolutional networks (GCN) into their network architectures, transforming an initial template to produce topologically accurate meshes \cite{kong2021deep,Ma2021,Bongratz2022,aase2023graph,Kruger2024}.

\hly{
}   

In this work, we propose a deep neural network-based method, i.e., KidMesh, which extracts computational meshes \cite{antiga2002geometric}  from MRU images, as shown in the right panel of Figure \ref{fig1}. 
\hly{
It is worth noting that our framework follows a similar paradigm to previous deep-learning-based mesh reconstruction methods such as Vox2Cortex~\cite{Bongratz2022}, which integrate volumetric feature extraction with graph-based mesh deformation. 
However, KidMesh is specifically tailored for pediatric CH and optimized to generate watertight and topologically consistent meshes that can be directly used for CFD simulation. 
} 
Our main contributions are:
\begin{itemize}
    \item \hly{We introduce KidMesh, the first end-to-end framework to generate simulation-ready meshes for pediatric CH directly from MRU scans.}
    \item \hly{A dynamic Vertex Upsampling (VU) module is proposed within a coarse-to-fine pipeline to handle the {heterogeneous geometry} of CH. This enables adaptive mesh refinement for accurately capturing  both large-scale structures and fine anatomical details.}
    \item \hly{We employ a weakly supervised strategy that learns from pseudo-gold standard meshes, eliminating the need for manual mesh-level annotations and enhancing clinical feasibility.}
    \item \hly{KidMesh achieves superior geometric accuracy and guarantees topological correctness, validated by successful CFD simulations that bridge the gap between anatomical structure and functional analysis.}
\end{itemize}

\section{Related Work}
\label{sec:Related Work}
\begin{figure}[htp]
    \centering
    \includegraphics[width=0.95\columnwidth]{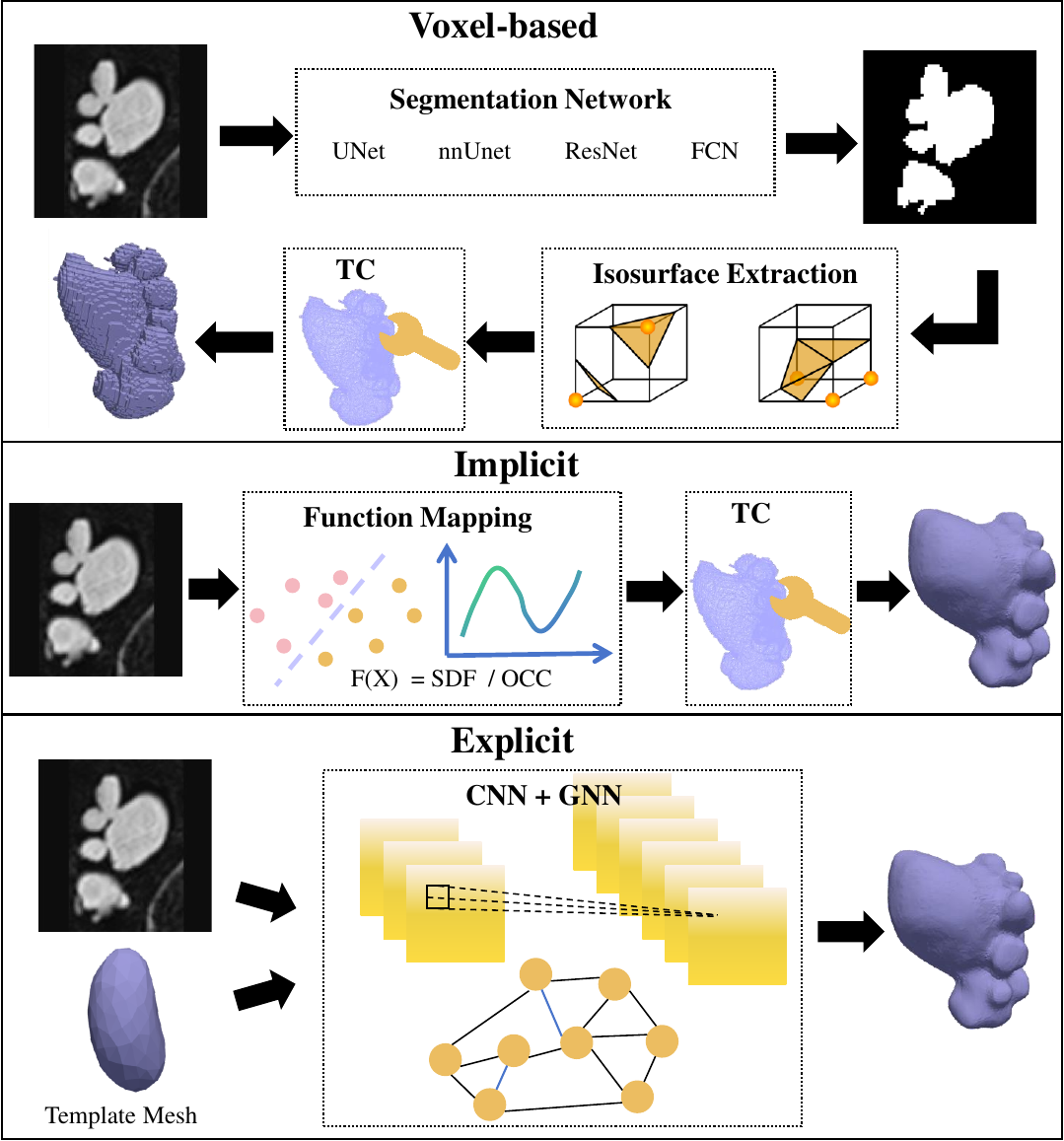}
    \caption{Workflow of the three mainstream deep learning mesh reconstruction methods: voxel-based, implicit, and explicit. The first two methods necessitate complex post-processing steps, such as isosurface extraction and topology correction (TC), following image segmentation (e.g., region-growing algorithms) or function mapping (e.g., signed distance function and occupancy function). In contrast, the explicit reconstruction method, which employs CNN and GCN, generates high-precision meshes while preserving the topology, thereby eliminating the need for post-processing.}
    \label{fig2}
\end{figure}

In recent years, deep learning has increasingly been employed for mesh reconstruction of target surfaces. 
Based on their principles, reconstruction methods can be categorized into three main types: voxel-based reconstruction, implicit reconstruction, and explicit reconstruction~\cite{Pan2019,Hu2021}. 
The basic principles of these methods are summarized in Figure~\ref{fig2}. 
To our knowledge, there is no research specifically addressing the reconstruction of CH meshes; we mainly review closely related works below.

\textbf{Voxel-based reconstruction.} 
Voxel-based mesh reconstruction methods utilize 3D voxel grids to generate surface mesh representations. 
Generally, these representations are achieved using isosurface extraction methods, such as the Marching Cubes algorithm~\cite{Lorensen1998a,Nielson2003,chen2024neural}. 
For instance, FastSurfer utilizes convolutional neural networks for 3D voxel grid segmentation. 
After that, it employs the Marching Cubes algorithm for surface generation and topology correction~\cite{Fischl2012a,Henschel2020}. 
The quality of the mesh generated by this method depends entirely on the resolution of the medical images, which is often determined by the voxel spacing of medical imaging protocols. 
Consequently, meshes derived from lower-resolution images exhibit staircase artifacts, leading to decreased geometric quality.

\textbf{Implicit reconstruction.} 
Implicit reconstruction methods overcome the resolution limitations of voxel-based approaches by learning continuous implicit functions that define surfaces as level sets of scalar fields rather than by predicting vertices directly. 
Common representations include occupancy fields~\cite{Mescheder2019}, signed distance functions (SDFs)~\cite{Park2019a,Xu2019}, and 3D Gaussian fields~\cite{Genova2019}. 
\hly{
In these methods, the final mesh vertices are extracted from the learned implicit fields using algorithms such as Marching Cubes, which enables the generation of smooth and high-resolution surfaces independent of voxel spacing. 
Building upon this idea, Santa Cruz et al.\ introduced DeepCSR, a geometric deep learning framework for cortical surface reconstruction~\cite{Cruz2021}. 
CortexODE~\cite{Ma2022CortexODE} further extended this line of work by modeling vertex evolution as continuous neural ordinary differential equation trajectories, representing a hybrid implicit formulation that learns continuous deformation fields through neural dynamics. 
More recently, Bongratz et al.\ proposed Neural Deformation Fields~\cite{Bongratz2024}, a template-based method that achieves accurate cortical shape recovery without dense supervision. 
Although these approaches have significantly advanced brain-surface modeling, they remain focused on cortical and white-matter geometries and have not been extended to CFD-ready anatomical meshes in other organs.}


\textbf{Explicit reconstruction.} 
Explicit reconstruction methods generate target meshes from template meshes by learning vertex displacements. 
For example, Wang et al.\ proposed a GCN that predicts vertex deformations for a spherical mesh~\cite{Wang2018a}, and Wickramasinghe et al.\ extended this concept to reconstruct smooth anatomical meshes such as the liver and hippocampus using topology-regularized loss functions~\cite{Wickramasinghe2020}. 
\hly{
In the cortical domain, CorticalFlow~\cite{Lebrat2021} and its extended version CorticalFlow++~\cite{CorticalFlowPP2022} employ diffeomorphic mesh transformers that explicitly deform a template mesh via smooth, topology-consistent transformations. 
These methods belong to continuous deformation-based explicit frameworks rather than purely implicit representations.}
For more complex structures like the cortex, Ma et al.\ deformed an initial white-matter mesh into the cortical surface using graph convolutions but relied on FreeSurfer preprocessing, which limits full automation~\cite{Ma2021}. 
Compared with these previous template-based methods (for example, MeshDeformNet~\cite{kong2021deep}), KidMesh introduces several important improvements.
It employs a learnable feature sampling module that adaptively aggregates MRU features through trilinear interpolation and cross-attention. 
It integrates topology-preserving and CFD-oriented regularization to maintain watertightness and geometric smoothness. 
It also leverages weak supervision from pseudo–gold standard meshes refined by radiologists, which allows its application to real clinical data without dense mesh annotations. 
These design choices enable anatomically faithful and simulation-ready reconstructions, particularly for hydronephrotic renal structures.

\textbf{Explicit reconstruction.} 
Explicit reconstruction methods generate target meshes from template meshes by learning vertex displacements. 
For example, Wang et al.\ proposed a GCN that predicts vertex deformations for a spherical mesh~\cite{Wang2018a}, and Wickramasinghe et al.\ extended this concept to reconstruct smooth anatomical meshes such as the liver and hippocampus using topology-regularized loss functions~\cite{Wickramasinghe2020}. 
In the cortical domain, CorticalFlow~\cite{Lebrat2021} and its extended version CorticalFlow++~\cite{CorticalFlowPP2022} employ diffeomorphic mesh transformers that explicitly deform a template mesh via smooth, topology-consistent transformations. 
These methods belong to continuous deformation-based explicit frameworks rather than purely implicit representations.
For more complex structures like the cortex, Ma et al.\ deformed an initial white-matter mesh into the cortical surface using graph convolutions but relied on FreeSurfer preprocessing, which limits full automation~\cite{Ma2021}. 
Compared with these previous template-based methods (e.g., MeshDeformNet~\cite{kong2021deep}), KidMesh introduces several important improvements. 
It employs a learnable feature sampling module that adaptively aggregates MRU features through trilinear interpolation and cross-attention. 
It integrates topology-preserving and CFD-oriented regularization to maintain watertightness and geometric smoothness. 
\hlb{While prior explicit reconstruction methods may also derive target meshes from volumetric masks \cite{kong2021deep}, they typically rely on post-processing to refine pseudo meshes for geometric supervision. In contrast, KidMesh is trained using pseudo-gold standard meshes extracted directly from MRU masks via Marching Cubes. These pseudo-gold standard meshes retain staircase artifacts and geometric noise, and KidMesh eliminates the need for manual mesh refinement through the proposed regularization terms.} 
These design choices enable anatomically faithful and simulation-ready reconstructions, particularly for hydronephrotic renal structures.

\hly{
In addition, recent explicit approaches have explored topology-aware deformation for biomechanical modeling. 
Examples include the distortion-energy formulation proposed by Pak et al.~\cite{pak2021distortion} for aortic valve mesh generation and the LinFlo-Net framework developed by Narayanan et al.~\cite{narayanan2024linflo} for producing simulation-ready cardiac meshes.
}
These studies share the objective of ensuring topological correctness and simulation compatibility. 
KidMesh extends these principles to fluid-filled renal anatomy and incorporates CFD-specific regularization to maintain surface continuity and watertightness for accurate urodynamic flow analysis.

\section{Method}
\label{sec:Method}
KidMesh is a deep-learning based explicit method, which integrates CNN and GCN to reconstruct the surface mesh of CH regions from MRU images. Figure \ref{fig3} illustrates the overall architecture of KidMesh, which consists of three main modules: feature extraction module (FEM),  feature sampling module (FSM), and  mesh deformation module (MDM). Specifically, KidMesh takes MRU images and a kidney-shaped ellipsoid mesh (template mesh) as inputs. After a series of deformations guided by image features, it outputs computational meshes suitable for CFD simulations. In the subsequent sections, we will provide the details of FEM (Section \ref{sec:feature extraction module}), FSM (Section \ref{sec:Feature Mapping Module}), MDM (Section \ref{sec:Mesh Deformation Module}), template mesh (Section \ref{sec:Mesh Template}), and loss functions (Section \ref{sec:Loss Function}).

\subsection{Feature Extraction Module (FEM)}
\label{sec:feature extraction module}
\begin{figure*}[htbp]
    \centering
    \includegraphics[width=0.9\textwidth]{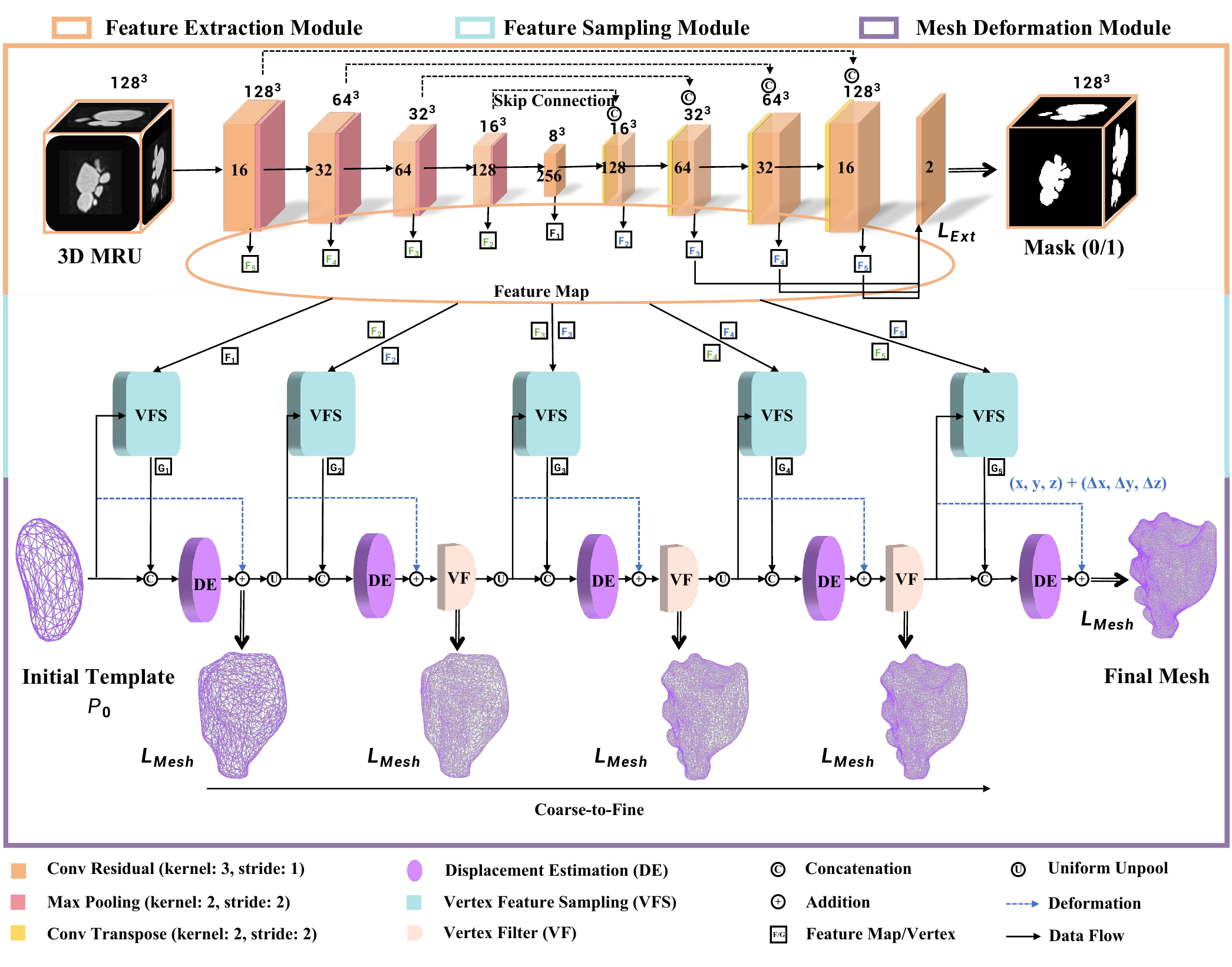}
    \caption{\hly{
    Overall architecture of KidMesh. 
    The network takes MRU images and a mesh template as input and generates computational meshes for congenital hydronephrosis in a coarse-to-fine manner. 
    The orange box denotes the Feature Extraction Module (FEM), which follows a U-shaped encoder–decoder topology to extract hierarchical 3D feature maps. 
    The blue box represents the Feature Sampling Module (FSM), which samples vertex-wise features via grid sampling. 
    The \hly{purple} box depicts the Mesh Deformation Module (MDM), which progressively deforms the mesh template into the target shape under multi-scale feature guidance.}}
    \label{fig3}
\end{figure*}

FEM is designed to extract representative feature maps from MRU images, which are subsequently propagated for mesh deformation. The upper panel of Figure \ref{fig3} illustrates FEM, \hlb{a 3D U-Net-like architecture that consists of an image encoder and decoder.} Let \(I\) denote an MRU image. Both the encoder and decoder employ a series of 3D residual convolutional layers to capture features at multiple levels:
\begin{equation}
    \{F^{}_1, F_2,\cdots F_L\} = \mathcal{F}_{\theta_{FE}}(I),
\end{equation}
where \(\{F_i | i = 1 \cdots L\}\) are the extracted feature maps from \(I\), and \(\theta_{FE}\) represents the learnable parameters of FEM.
\hly{
FEM contains four encoder–decoder blocks as well as one bottleneck block. Each block includes two 3D convolutional layers with a kernel size of \(3\times3\times3\) and a stride of \(1\times1\times1\), followed by batch normalization and ReLU activation. 
Neighboring blocks are connected by max-pooling layers with a stride of \(2\times2\times2\) in the encoder and transposed convolutions in the decoder. 
Residual shortcuts are introduced within each block to stabilize gradient propagation, and skip connections link the encoder and decoder blocks for multi-scale fusion.}

\hly{The input MRU image has dimensions of \(128^3\). 
FEM produces five levels of feature maps \(\{F_1, F_2, F_3, F_4, F_5\}\) with spatial resolutions of  \(8^3\), \(16^3\), \(64^3\), \(32^3\) and \(128^3\),  and channel dimensions of 256, 128,  64, 32,  and 16. 
These multi-scale features collectively encode both geometric and contextual information for the downstream FSM.
}

\hly{
We train FEM to generate CH binary segmentation masks of the input image, guiding the network to focus on boundary-sensitive features that are critical for identifying hydronephrosis. 
To improve training stability and gradient flow, a deep supervision strategy is adopted, where intermediate decoder outputs are directly supervised across multiple scales~\cite{lee2015deeply,wang2015training,zhang2018deep}. 
This design helps FEM preserve spatially detailed representations and enhances the quality of the feature maps delivered to the mesh reconstruction module.
}

\subsection{Feature Sampling Module (FSM)}
\label{sec:Feature Mapping Module}
\hly{FSM serves as the bridge between FEM and MDM. It computes detailed feature representations for each mesh vertex by performing grid sampling and graph-based feature aggregation. 
The middle panel of Figure~\ref{fig3} illustrates the architecture of FSM. 
The module first applies graph convolutions to predict per-vertex offset vectors that represent small 3D displacements around each vertex position. 
These offsets specify a set of learnable sampling points in the local vicinity of the vertex, thereby defining its adaptive neighborhood range for feature extraction. 
Volumetric features from FEM are then interpolated at these offset locations through grid sampling, and a sequence of GCN layers further aggregates and refines the sampled vertex features:}
\begin{equation}
    G_i = \mathcal{F}_{\theta_{FS}}[GS(F^{}_i, P_{i-1})],
\end{equation}
where \(G^{}_{i}\) denotes the sampled feature vertex, \(GS(\cdot)\) is the grid sampling operation, \(P_{i-1}\) represents the mesh at step \(i - 1\), and \(\theta_{FS}\) indicates the learnable parameters of FSM.  
\hly{
The term “feature vertices” refers to vertex-wise feature descriptors obtained by sampling the 3D MRU feature maps onto mesh vertices through the grid sampling process. 
These descriptors capture local contextual information around each vertex and provide the feature input for subsequent mesh deformation stages. 
}

\hly{
In practice, the grid sampling operation \(\mathrm{GS}(\cdot)\) applies trilinear interpolation to project 3D feature maps \(\mathbf{F}_i\) onto mesh vertices. 
Each vertex with normalized coordinates \((x, y, z) \in [-1,1]^3\) queries its corresponding feature using \texttt{grid\_sample} (mode=``bilinear", align\_corners=``True", padding\_mode="border"). 
A learnable offset field \(\Delta \mathbf{s}\) further defines a \(3\times3\times3\) local neighborhood, from which nearby features are re-sampled and aggregated through a small convolutional kernel. 
This mechanism ensures smooth and spatially} \hly{consistent feature propagation from MRU volumes to mesh vertices.
}

\hly{
After local aggregation, FSM enhances global geometric coherence through a lightweight self-attention mechanism across} \hly{mesh vertices. 
Unlike traditional cross-attention that operates between different modalities, this self-attention models long-range dependencies within the same set of vertex features. 
We compute a normalized pairwise correlation matrix
\(\mathbf{A}_{mn} = \frac{\mathbf{f}_m \cdot \mathbf{f}_n}{\|\mathbf{f}_m\|\|\mathbf{f}_n\|}\),
where \(\mathbf{f}_m\) and \(\mathbf{f}_n\) are vertex features at indices \(m\) and \(n\). 
The updated feature of vertex \(m\) is obtained as 
\(\mathbf{g}_m = \sum_n \mathbf{A}_{mn}\mathbf{f}_n\),
allowing information exchange among spatially distant but anatomically related vertices. 
This parameter-free self-attention across mesh vertices complements local graph aggregation, improving global smoothness and topological consistency.
}

During the early stages of mesh deformation, FSM propagates high-level abstract feature maps that capture global structural patterns and guide coarse displacement estimation. As deformation progresses, the module gradually transitions to lower-level feature maps with higher spatial resolution, enabling fine-grained reconstruction. Through this coarse-to-fine refinement process, KidMesh produces anatomically consistent and high-quality 3D meshes of CH regions.

\subsection{Mesh Deformation Module (MDM)}
\label{sec:Mesh Deformation Module}
MDM is designed to predict vertex displacements for progressive mesh deformation. 
Starting from a template mesh, this module incrementally deforms it to match the target CH regions while preserving topological consistency. 
The bottom panel of Figure~\ref{fig3} illustrates the overall architecture of the deformation module. 
It consists of several displacement estimation (DE) stages, where each stage predicts vertex-wise displacements \((\Delta x, \Delta y, \Delta z)\) for the input mesh. 
Each DE step takes as input the current mesh and its corresponding vertex features to estimate displacement fields:
\begin{equation}
    \label{eq3}
    \Delta {\Phi}_{i+1} = \mathcal{F}_{\theta_{DE}}(G_{i+1}, {P}_{i}),
\end{equation}
where \(\Delta \Phi_{i+1}\) denotes the predicted displacement, \(G_{i+1}\) represents feature vertex derived from FSM, \({P}_{i}\) is the mesh predicted by the \(i\)-th DE step, and \(\theta_{DE}\) denotes the learnable parameters of MDM. 
Here, \(P_0\) is the initial template mesh, and the deformed mesh is updated as:
\begin{equation}
    {P}_{i+1} = {P}_{i} + \Delta \Phi_{i+1}.
\end{equation}

\hly{
Each deformation stage in MDM is composed of three GraphConv layers followed by Batch Normalization and ReLU activation, 
and a final linear layer that outputs vertex displacements \(\Delta \Phi_i \in \mathbb{R}^{N\times3}\). 
Residual connections between stages are maintained to ensure stable gradient propagation and smooth deformation. 
Each stage shares the same architectural design but differs in feature dimensionality, which gradually decreases from 256 channels at the coarsest level to 16 at the finest level, mirroring the hierarchical features extracted by FEM.
}

\begin{figure}[htbp]
    \centering
    \includegraphics[width=0.8\columnwidth]{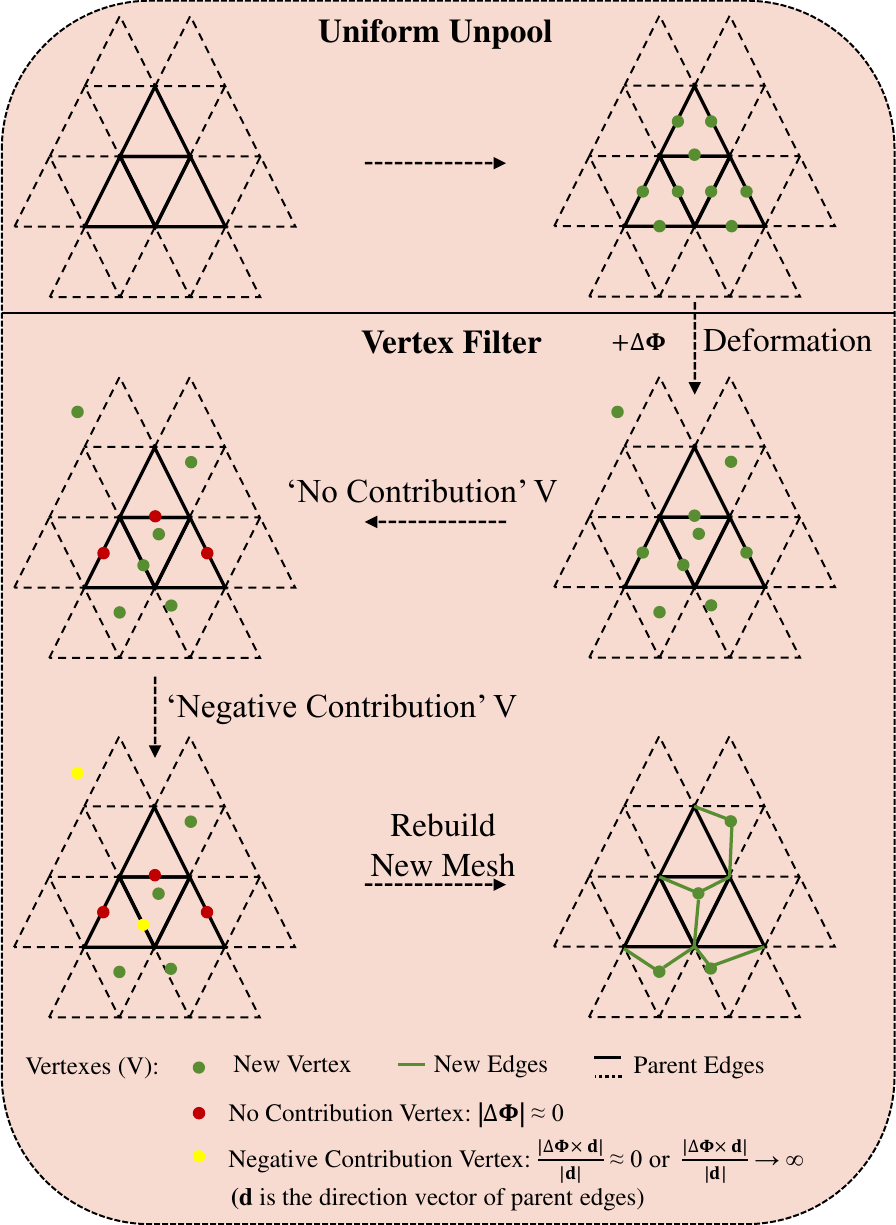}
    \caption{The illustration of vertex upsampling, consists of uniform unpool (UU) and vertex filter (VF). UU aims to enlarge the number of vertex in generated meshes, while VF filters out no and negative contribution vertices.}
    \label{cropped_Figure_sup3}
\end{figure}

As deformation proceeds, KidMesh employs a VU strategy to enhance reconstruction accuracy. 
The mesh resolution is progressively increased through uniform unpool (UU) operations, which enable fine reconstruction of CH regions with complex shapes or strong local variations. 
The upper panel of Figure~\ref{cropped_Figure_sup3} illustrates the UU process, where the midpoint of each edge is generated to subdivide a triangular face into four smaller ones~\cite{Balin2023}. 
Some newly generated vertices, however, may contribute little or even negatively to the reconstruction. 
To address this issue, a vertex filter (VF) is introduced to selectively prune such vertices based on their geometric contribution, as shown in the lower panel of Figure~\ref{cropped_Figure_sup3}. 
Vertices that remain stationary or show excessive displacement relative to their parent edges are removed. 
Through this mechanism, VF allows KidMesh to adaptively refine mesh resolution while maintaining geometric stability.
\hlb{
Furthermore}, the vertex upsampling and filtering steps are executed dynamically during training rather than being precomputed. Between successive deformation stages, vertex subdivisions are generated on the fly according to the topological structure of the base mesh, which ensures consistent adjacency and smooth connectivity. 

\subsection{Template Mesh}
\label{sec:Mesh Template}
To simplify the deformation process, we assume that CH regions are closed spaces with a single connected component. 
Therefore, any mesh with spherical topology (\(g = 0\)) can be theoretically used as input. 
\hly{
In this work, the template meshes were programmatically generated rather than derived from real MRU scans. 
A smooth implicit surface was first fitted to the averaged boundary of the hydronephrosis masks in the training set to create a canonical kidney-shaped reference. 
This surface was then converted into a watertight triangular mesh using the Marching Cubes algorithm, and isotropic remeshing was applied to achieve uniform vertex distribution and stable topology. 
To support multi-resolution deformation, three template} \hly{meshes were generated through spherical subdivision, containing 56, 162, and 642 vertices, respectively. 
All templates were normalized to the \([-1, 1]^3\) coordinate range and consistently oriented across samples.
}
The resulting kidney-shaped templates\footnote{\url{https://github.com/Mrdeer3935/KidMesh}} improve reconstruction accuracy and provide a stable initialization for deformation. 
\hly{
Adjusting the number of vertices offers a practical balance between reconstruction precision and training efficiency, 
allowing KidMesh to adapt to different levels of anatomical detail.
}

\subsection{Loss Function}
\label{sec:Loss Function}
KidMesh is trained by a hybrid loss function, which consists of extraction ($\mathcal{L}_{Ext}$) and reconstruction ($\mathcal{L}_{Rec}$) losses. The extraction loss combines cross-entropy loss and Dice loss to ensure accurate feature map extraction. The reconstruction loss consists of a geometric loss (Chamfer Loss) and several regularization terms, including Laplacian loss, normal loss, edge length loss, self-intersection loss, and area loss. These terms help optimize MDM, enabling it to deform a template mesh to match target CH regions.

\subsubsection{Extraction Loss}
\label{sec:Segmentation Loss}
KidMesh adopts a loss function, which includes Dice loss and cross-entropy loss, to train FEM. Let $\hat{M}$ denote the predicted binary CH masks, and $M$ is the gold standard CH masks.

\textbf{Dice Loss:} The Dice loss function is used to measure the similarity between the predicted and the gold standard masks as follows:
\begin{equation}
    \mathcal{L}_{Dice}\left( {\hat{M},M} \right) = 1 - \frac{2 \times \left| {\hat{M} \cap M} \right|}{\left| \hat{M} \right| + \left| M \right|},	
\end{equation}
where $\left| {~ \cdot ~} \right|$ calculates the number of elements in a set, and $\cap$ denotes the intersection between sets.

\textbf{Cross-Entropy Loss:} The cross-entropy loss function is used to measure the log-likelihood between the predicted and the gold standard masks. It is defined as follows:
\begin{equation}
    \begin{aligned}
        \mathcal{L}_{cross-entropy}&\left( {\hat{M},M} \right) = - \sum_{c \in \Omega} \Big[ M(c) \log\left( {\hat{M}(c)} \right) \\
        & + \left( 1 - M(c) \right) \log\left( {1 - \hat{M}(c)} \right) \Big],
    \end{aligned}
\end{equation}
where $c$ is a spatial point in \hly{image space} $\Omega$.

\subsubsection{Reconstruction Loss}
\label{sec:Reconstruction Loss}
Measuring reconstruction similarity between two meshes is challenging. Firstly, obtaining a gold standard mesh for the CH region is highly difficult. Secondly, this difficulty arises from the lack of a one-to-one correspondence between mesh vertices due to potential differences in vertex count. 
\hly{
To address this issue, we generated pseudo–gold standard meshes from CH masks using the Marching Cubes algorithm and trained KidMesh under a weakly supervised setting.  
In this context, “weak supervision” refers to the use of pseudo meshes that are automatically produced without manual annotation or vertex-level correspondence, serving as geometric supervision during training.  
Rather than relying on direct vertex-to-vertex matching, the training process is guided by differentiable distance-based metrics such as Chamfer and Normal losses, allowing the network to learn topology-preserving deformations even when supervision is imperfect.  
No additional smoothing was applied to the pseudo meshes, as they were intentionally kept with staircase artifacts to demonstrate that KidMesh can learn a self-corrective geometric prior and reconstruct smooth, watertight surfaces from imperfect reference data.
}
Meanwhile, we \hly{incorporated} multiple regularization terms to ensure the topology of reconstructed 3D meshes.  
The loss functions are as follows:

\textbf{Chamfer Loss:} The Chamfer loss to measure the geometric shape differences between the predicted mesh $P$ and the pseudo gold standard mesh $Q$. Specifically, the approximation between the two mesh point sets is evaluated. The Chamfer loss is bidirectional, meaning that it calculates the minimum distance from each point in the predicted mesh point set to the pseudo gold standard mesh point set, and vice versa:
\begin{equation}
    \begin{aligned}
        \mathcal{L}_{chamfer}\left( {P,Q} \right) = & \sum_{p \in P} \min_{q \in Q} \parallel p - q \parallel^{2} \\
        & + \sum_{q \in Q} \min_{p \in P} \parallel q - p \parallel^{2},
    \end{aligned}
\end{equation}
where  $P$ and $Q$ are predicted mesh and pseudo-gold-standard mesh, respectively.

\textbf{Laplacian Loss:} The Laplacian loss aims to maintain the smoothness of the mesh surface and reduce irregularities. KidMesh promotes an even distribution of the mesh surface by minimizing the average distance between each mesh vertex and its neighboring vertices:
\begin{equation}
    \mathcal{L}_{Laplacian}(P) = \sum_{p \in P}\,~{\parallel {p - \frac{1}{|\mathcal{N}(p)|}\sum_{c \in \mathcal{N}(p)}\,c} \parallel}^{2},
\end{equation}
where $\mathcal{N}(p)$ is the set of neighboring vertices of vertex $p$, and $|\cdot|$ denotes the number of elements in $\mathcal{N}(p)$.

\textbf{Normal Loss:} The normal loss minimizes the difference between the predicted and pseudo gold standard mesh normals, promoting the normal directions to be as consistent (parallel) as possible:
\begin{equation}
    \begin{aligned}
        \mathcal{L}_{normal}(P) = & \sum_{p \in P} \sum_{\substack{q = \arg\min_{q} \\ (\parallel p - q \parallel^{2})}} \parallel \langle (p_{1} - p) \times (p_{2} - p), n_{q} \rangle \parallel^{2},
    \end{aligned}
\end{equation}
where $p_{1}$ and $p_{2}$ are the two vertices that share the same face with vertex $p$, $n_{q}$ is the normal of vertex $q$ in the pseudo gold standard mesh $Q$. $\times$ and $ < \cdot ~,~ \cdot > $ are vector cross product and dot product, respectively. Here, the normal of each triangular face is calculated by the cross product of two edges (i.e., ${p_{1} - p}$ and ${p_{2} - p}$) on the face.

\textbf{Edge Length Loss:} The edge length loss is used to constrain the length of mesh edges, ensuring that the generated mesh maintains appropriate detail and shape. This loss function \hlb{controls} the length of mesh edges to reduce the occurrence of long edges, and thereby preventing mesh degradation:
\begin{equation}
    \mathcal{L}_{edge}(P) = \sum_{p \in P}~\,\sum_{c \in \mathcal{N}(p)}\,{\parallel p - c \parallel}^{2}.
\end{equation}
\hly{
where \(c \in N(p)\) denotes a vertex directly connected to \(p\) by an edge, i.e., a one-ring neighbor defined by the mesh connectivity.  
The adjacency originates from the initial template and is consistently expanded through deterministic upsampling operations during deformation.  
Within each stage, it remains fixed to preserve local topology while newly added vertices inherit the connectivity of their parent edges.}

\textbf{Area Loss:} 
The area loss function aims to constrain triangle sizes within  meshes. Here, the area of each triangle is calculated using the cross product of vectors, and the average of these areas is used as a single scalar value to quantify the uniformity of the mesh:
\begin{equation}
    \mathcal{L}_{area }(P)=\frac{1}{|F|} \sum_{\left\{p_1, p_2, p\right\} \in f, f \in F} \frac{1}{2}\left\|\left(p_1-p\right) \times\left(p_2-p\right)\right\|^2,
\end{equation}
where $F$ is the set of all faces in the predicted mesh $P$, and $p_{1},~p_{2},~p$ denotes the vertices of face $f$. 

\textbf{Seal Loss:}  
\hlb{
The Seal Loss is a differentiable regularization term designed to suppress surface validity failures during mesh deformation. Since mesh connectivity is preserved throughout deformation and upsampling, watertightness cannot be enforced via discrete topological tests. Instead, we introduce a geometric surrogate that penalizes effective loss of surface closure caused by geometric degeneracies, following common practices in geometric processing~\cite{botsch2010polygon}.}
\begin{equation}
\mathcal{L}_{seal}(P) = \frac{1}{|E|}\sum_{e\in E}(1-\delta_{e,2})
+ \lambda\sum_{(f_i,f_j)\in \mathcal{N}(f_i)}(1-n_i\cdot n_j),
\end{equation}
where \(E\) denotes the set of mesh edges. The term \(\delta_{e,2}\) is a continuous geometric indicator derived from the areas of the two faces incident to an edge, measuring whether the edge is effectively supported by two non-degenerate faces. The second term enforces local normal consistency between adjacent faces with normals \(n_i\) and \(n_j\).  
\hlb{Together, these terms penalize geometric failure modes such as face area collapse, surface folding, or normal inversion, providing a differentiable approximation of surface closure that improves geometric regularity and robustness during training.
}

\subsubsection{Total Loss}
\label{sec:Total Loss}
Finally, KidMesh employs a multi-scale training strategy. Specifically, we utilize the output of the last three layers of FEM, and resample them to a uniform size for network training:
\begin{equation}
    \begin{aligned}
        \mathcal{L}_{Ext} = & \sum_{i \in \{1,2,3\}} \gamma_i \Big[ \rho_1 \mathcal{L}_{Dice}\left(\hat{M}_i, M\right) \\
        & + \rho_2 \mathcal{L}_{cross-entropy}\left(\hat{M}_i, M\right) \Big],
    \end{aligned}
\end{equation}
where $\hat{M}_{i}$ is the resized output of the $i$-th layer of the decoder of FEM, and $\gamma_{i}$ is a  hyper-parameter. $\rho_{1}$ and $\rho_{2}$ are the weights for the Dice loss and cross-entropy loss, respectively.

The reconstruction loss is computed at each DE step, each comprises weighted terms as follows:
\begin{equation}
    \begin{aligned}
        \mathcal{L}_{Rec} = &\sum_{i \in\{1,2,3,4,5\}} \left[ \alpha_1 \mathcal{L}_{chamfer}\left(P_i, Q\right) \right. \\
        &+ \alpha_2 \mathcal{L}_{Laplacian}\left(P_i\right) + \alpha_3 \mathcal{L}_{normal}\left(P_i\right) \\
        &+ \alpha_4 \mathcal{L}_{edge}\left(P_i\right) + \alpha_5 \mathcal{L}_{area}\left(P_i\right) \\
        &\left. + \alpha_6 \mathcal{L}_{seal}\left(P_i\right) \right],
    \end{aligned}
\end{equation}
where \(P_{i}\) denotes the predicted mesh from the \(i\)-th DE step, and \(\alpha_{1\sim6}\) are the weight coefficients used to balance different loss terms. 

\hly{The entire KidMesh framework is trained in an end-to-end manner, in which the FEM, FSM, and MDM are jointly optimized under the hybrid loss \((\mathcal{L}_{Rec} + \mathcal{L}_{Ext})\).  
Such unified optimization allows gradients to propagate seamlessly across all components, ensuring consistent feature representations between voxel-, vertex-, and mesh-level spaces and leading to stable and coherent convergence of the network.}


\section{Experiments and Results}
\label{sec:Experiments and Results}

\subsection{Dataset and Preprocessing}
\label{sec:Dataset and Preprocessing}
The experimental dataset consists of MRU images collected from 160 pediatric patients, each containing both pre-operative and post-operative MRU scans. 
\hly{
All images were acquired on the same 3.0~T Siemens MAGNETOM Skyra scanner using a standardized T2-weighted 3D turbo spin-echo MRU protocol (slice thickness:~1.0~mm, field-of-view:~240~mm, matrix:~256$\times$256). 
This ensured consistent imaging parameters across all cases.
The dataset was randomly divided into 112, 32, and 16~cases for training, validation, and testing, respectively (70\%, 20\%, and 10\%)}. 
\hly{
Before further processing, all MRU volumes were reoriented to a standard radiological orientation following the RAS coordinate convention using the ITK toolkit. 
This ensured consistent anatomical alignment across subjects and scanning sessions.
}
Each MRU image was \hly{automatically center-cropped according to the centroid of the nnU-Net segmentation mask} to extract a region of interest (ROI) that contained the CH region. 
All volumes were resampled to an isotropic resolution of 1~mm~$\times$~1~mm~$\times$~1~mm, then resized to (128, 128, 128) and normalized using Z-score normalization~\cite{Cheadle2003}. 
\hly{
To improve generalization, random 3D augmentations were applied during training, including rotation (\(\pm15^\circ\)), scaling (0.9–1.1), Gaussian noise addition (\(\sigma=0.01\)), and elastic deformation (grid size=8, magnitude=5).
}

\hly{
KidMesh requires pseudo–gold standard meshes of all MRU images for both training and evaluation. 
CH masks were manually annotated by two board-certified pediatric radiologists with over five years of experience in urinary tract imaging. 
These expert-annotated volumetric segmentations served as the reference “gold standard” in this study. 
The corresponding pseudo–gold standard meshes were extracted from these manual masks using the Marching Cubes algorithm (isovalue=0.5, voxel spacing=1~mm), followed by post-processing steps such as hole filling and the removal of isolated components smaller than 50~voxels~\cite{Shen2023,Guillard2024}. 
No further smoothing was applied, as KidMesh is designed to reconstruct smooth and watertight surfaces directly from unsmoothed pseudo meshes that retain staircase-like characteristics. 
Finally, all vertex coordinates were normalized to the range of [–1, 1].}


\subsection{Implementation Details}
\label{sec:Experiment Details}
\hly{KidMesh was implemented in Python~(v3.11.8) using PyTorch~(v2.1.0) and PyTorch3D~(v0.7.5)~\cite{Paszke2019,Ravi2020}. The corresponding hyperparameters are listed in Table~\ref{tab:loss_weights}, and the overall training configuration is summarized in Table~\ref{tab:hyperparams_sup}.
To determine appropriate weighting factors for each loss component, a comprehensive grid search was conducted over predefined candidate ranges.  
For the extraction loss $\mathcal{L}_{Ext}$, the multi-scale weights $\gamma_{1,2,3}$ were explored within $\{0.05, 0.1, 0.3, 0.5, 0.9\}$, and the Dice-to-cross-entropy ratios $\rho_{1,2}$ within $\{(0.7,0.3), (0.5,0.5), (0.3,0.7)\}$. This follows common practices in medical image segmentation networks~\cite{isensee2018nnu,milletari2016v}.  
For the reconstruction loss $\mathcal{L}_{Rec}$, the coefficients $\alpha_{1\sim6}$ were searched across logarithmic scales $\{0.01, 0.1, 1.0, 10.0\}$, consistent with prior mesh reconstruction frameworks that balance geometric fidelity and surface regularization~\cite{Wickramasinghe2020,Cruz2021}.  
Each configuration was evaluated using validation metrics that jointly measure geometric accuracy (Chamfer and normal distances) and topological stability (self-intersection count and watertightness ratio).  
The final parameters $\alpha_{1\sim6}{=}\{1.0,0.1,0.1,0.1,1.0,0.1\}$ achieved the best trade-off between reconstruction accuracy and smoothness, in line with strategies adopted in other deformation-based medical mesh reconstruction studies~\cite{kong2022learning}.  
All experiments were carried out under identical conditions to ensure reproducibility.}

\subsection{Evaluation Metrics}
\label{sec:Evaluation Metrics}
We evaluated the reconstruction quality of KidMesh using average symmetric surface distance (ASSD), Hausdorff distance (HD), point-to-surface distance (P2SD), Dice, and Jaccard scores~\cite{Bongratz2023,Bongratz2024}. Specifically, ASSD and HD were computed between reconstructed meshes and pseudo–gold standard meshes derived from refined CH masks, denoted as ASSD$_\text{mesh}$ and HD$_\text{mesh}$. \hly{To further assess boundary localization accuracy, we also extracted voxel-level boundary points from the pseudo–gold standard masks and computed the same metrics, denoted as ASSD$_\text{outer}$ and HD$_\text{outer}$. While both originate from the same reference, the “mesh” metrics measure continuous geometric fidelity of the reconstructed surfaces, whereas the “outer” metrics emphasize voxel-level boundary alignment and localization precision in volumetric space. This dual evaluation provides complementary perspectives on global shape recovery and local contour accuracy. Moreover, P2SD quantifies fine-grained geometric deviation between the reconstructed and reference meshes, capturing subtle surface irregularities. Since KidMesh outputs 3D meshes rather than voxel masks, all reconstructed meshes were rasterized into voxel grids for fair comparison with voxel-based segmentation baselines.}

\begin{table*}[htbp]
    \centering
    \caption{Comparison of KidMesh performance with different configurations. "$_\text{mesh}$" and "$_\text{outer}$" indicate two different evaluation strategies based on pseudo gold standard mesh and outer surface point set, respectively. Dice and Jaccard scores were calculated after rasterizing meshes into voxel grids. VU: vertex upsampling. Mesh\_NO: number of vertex in template mesh.}
    \label{table3}
    \resizebox{\textwidth}{!}{
        \begin{tblr}{
                width = \linewidth,
                colspec = {Q[80]Q[80]Q[100]Q[100]Q[95]Q[95]Q[90]Q[120]Q[90]},
                cells = {c,m},
                hline{1-2,5,8} = {-}{},
            }
            \textbf{VU} & \textbf{Mesh\_NO} & \textbf{ASSD$_\text{outer}$↓} & \textbf{ASSD$_\text{mesh}$↓} & \textbf{HD$_\text{outer}$↓} & \textbf{HD$_\text{mesh}$↓} & \textbf{Dice(\%)↑} & \textbf{Jaccard(\%)↑} & \textbf{P2SD↓} \\
            \textbf{3} & \textbf{56} & 2.62±.192 & 2.56±.210 & 8.86±.289 & 8.80±.239 & 84.0±2.51 & 72.9±1.85 & 1.28±.213 \\
            \textbf{3} & \textbf{162} & 2.11±.256 & 2.04±.320 & 8.22±.576 & 8.16±.640 & 86.0±3.58 & 78.3±2.09 & 1.15±.192 \\
            \textbf{3} & \textbf{642} & 1.60±.121 & 1.66±.123 & 8.02±.223 & 8.00±.320 & 87.6±2.46 & 79.2±2.41 & 1.14±.256 \\
            \textbf{2} & \textbf{162} & 2.64±.289 & 2.62±.159 & 8.80±.408 & 8.68±.571 & 84.6±1.56 & 73.7±1.84 & 1.21±.145 \\
            \textbf{3} & \textbf{162} & 2.11±.256 & 2.04±.320 & 8.22±.576 & 8.16±.640 & 86.0±3.58 & 78.3±2.09 & 1.15±.192 \\
            \textbf{4} & \textbf{162} & 1.66±.134 & 1.53±.115 & 7.89±.234 & 7.82±.331 & 88.8±1.50 & 79.5±1.81 & 1.02±.123 \\
        \end{tblr}
    }
\end{table*}

\subsection{Configuration Parameter Study}
\label{sec:Model Flexibility}
\begin{figure*}[htbp]
    \centering
    \includegraphics[width=0.8\textwidth]{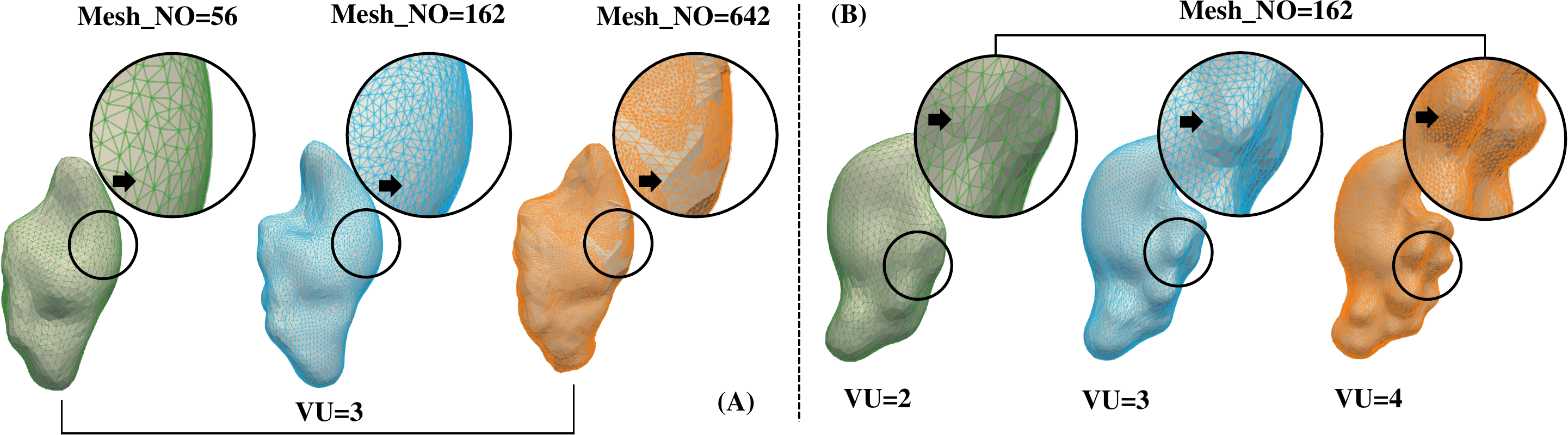}
    \caption{Visualization of the reconstructed meshes of KidMesh with different configurations. Black circles highlight the regions where have significant mesh differences.}
    \label{fig6}
\end{figure*}

We first investigated the optimal configuration of KidMesh. In particular, the resolution (number of vertices) of input meshes and the number of VU steps play crucial roles in determining its performance. As listed in the upper panel of Table \ref{table3},  reconstruction performance improved with larger resolution of meshes. However, excessively increasing resolution could lead to uneven vertex distribution or self-intersecting surfaces, as indicated by the arrows in Figure \ref{fig6}(A). Therefore, we selected an initial mesh with 162 vertices as the KidMesh template.

Additionally, reconstruction performance could be further improved by increasing the number of VU steps, as shown in the lower panel of Table \ref{table3}. In Figure \ref{fig6} (B), KidMesh generated more plausible reconstruction details in the renal calyx regions, as indicated by the arrows. However, increasing the VU steps led to a significant rise in computational resource demands. Specifically, GPU memory usage and inference time increased by a factor of 28 and 13, respectively, when the VU step was increased from 3 to 4. Therefore, we set the VU steps of KidMesh to 3 to strike a balance between performance and resource consumption.

\begin{table}[htbp]
    \centering
    \caption{Computational resource consumption of KidMesh with different configurations. FLOPS: Floating-point Operations Per Second, G: Giga, GB: Gigabyte, S: second.}
    \label{table2}
    \resizebox{\columnwidth}{!}{
        \begin{tblr}{
                width = \linewidth,
                colspec = {Q[117]Q[192]Q[127]Q[187]Q[165]},
                cells = {c,m},
                hline{1-2,5,8} = {-}{},
            }
            \textbf{VU} & \textbf{Mesh\_NO} & \textbf{FLOPS (G)} & \textbf{GPU Mem (GB)} & \textbf{Infer Time (s)} \\
            \textbf{3} & \textbf{56} & 224.60 & 3.34 & 0.12 \\
            \textbf{3} & \textbf{162} & 2545.65 & 5.89 & 0.36 \\  
            \textbf{3} & \textbf{642} & 154290.77 & 36.97 & 13.05 \\
            \textbf{2} & \textbf{162} & 255.31 & 3.79 & 0.11 \\
            \textbf{3} & \textbf{162} & 2545.65 & 5.89 & 0.36 \\  
            \textbf{4} & \textbf{162} & 71481.71 & 21.41 & 5.93 \\   
        \end{tblr}
    }
\end{table}

\subsection{Comparison Study}
\label{sec:Comparative Results}
We compared KidMesh with existing mesh reconstruction methods. It is important to note that voxel-based segmentation methods require manual or semi-automatic topological correction and post-processing procedures to reconstruct meshes. However, KidMesh could produce accurate and smooth meshes without the need for extensive procedures. 
\begin{itemize}
    \item nnU-Net: nnU-Net is a well-established model for medical image segmentation, and has been successfully applied to various segmentation tasks \cite{Ronneberger2015}. 
    \item Fully Convolutional Network (FCN): FCN is the first semantic segmentation model based on fully convolution layers \cite{Roth2018}. 
    \item Spatial and Channel Attention-enhanced U-Net (SCU): SCU is a variant of U-Net that incorporates spatial and channel attention mechanisms to enhance segmentation accuracy \cite{Gohil2021}. This method was first utilized in the KiTS21 Challenge, hosted by MICCAI, which demonstrated promising performance on kidney segmentation.
    \item  ResNet-50 (RES50): RES50 is a segmentation model based on a pre-trained model with residual structure. Here, the pre-trained model was trained by using large-scale medical datasets across different modalities, organs, and pathologies \cite{Chen2019}.
    \item MeshDeformNet: MeshDeformNet is a state-of-the-art (SOTA) whole-heart mesh reconstruction method. It is widely applied in cardiac research \cite{kong2021deep}.
\end{itemize}


\hly{
For a fair comparison, all voxel-based baselines (nnU-Net, SCU, RES50, and FCN) were subjected to identical post-processing before quantitative evaluation. 
Specifically, only the largest connected component of each predicted mask was retained to remove small spurious fragments, followed by 3D morphological opening and closing to eliminate isolated clusters and fill minor holes. 
A Gaussian smoothing filter was then applied to reduce voxel-level noise prior to mesh extraction using the Marching Cubes algorithm. 
These standardized steps suppressed outliers that could affect Hausdorff Distance while maintaining the anatomical integrity of the CH regions.
}

\hly{
After rasterizing the meshes generated by KidMesh, it achieved higher Dice and Jaccard scores than FCN and SCU, while showing slightly lower accuracy than nnU-Net and RES50, as presented in Table~\ref{table1}. 
When the segmentation masks from voxel-based methods were converted into meshes, KidMesh demonstrated markedly better results in terms of HD and P2SD across all segmentation-based approaches. 
These findings indicate that voxel-based methods remain effective for CH region segmentation, yet they tend to produce outliers and staircase artifacts more frequently than KidMesh. 
In contrast, KidMesh significantly reduced reconstruction time, completing mesh generation in an average of 0.36~seconds, whereas RES50 combined with post-processing required 86.5~seconds. 
Compared with the state-of-the-art mesh deformation model MeshDeformNet, KidMesh also achieved superior results in all geometric and topological metrics, demonstrating both stability and accuracy.} 

\begin{figure}[htbp]
    \centering
    \includegraphics[width=0.98\columnwidth]{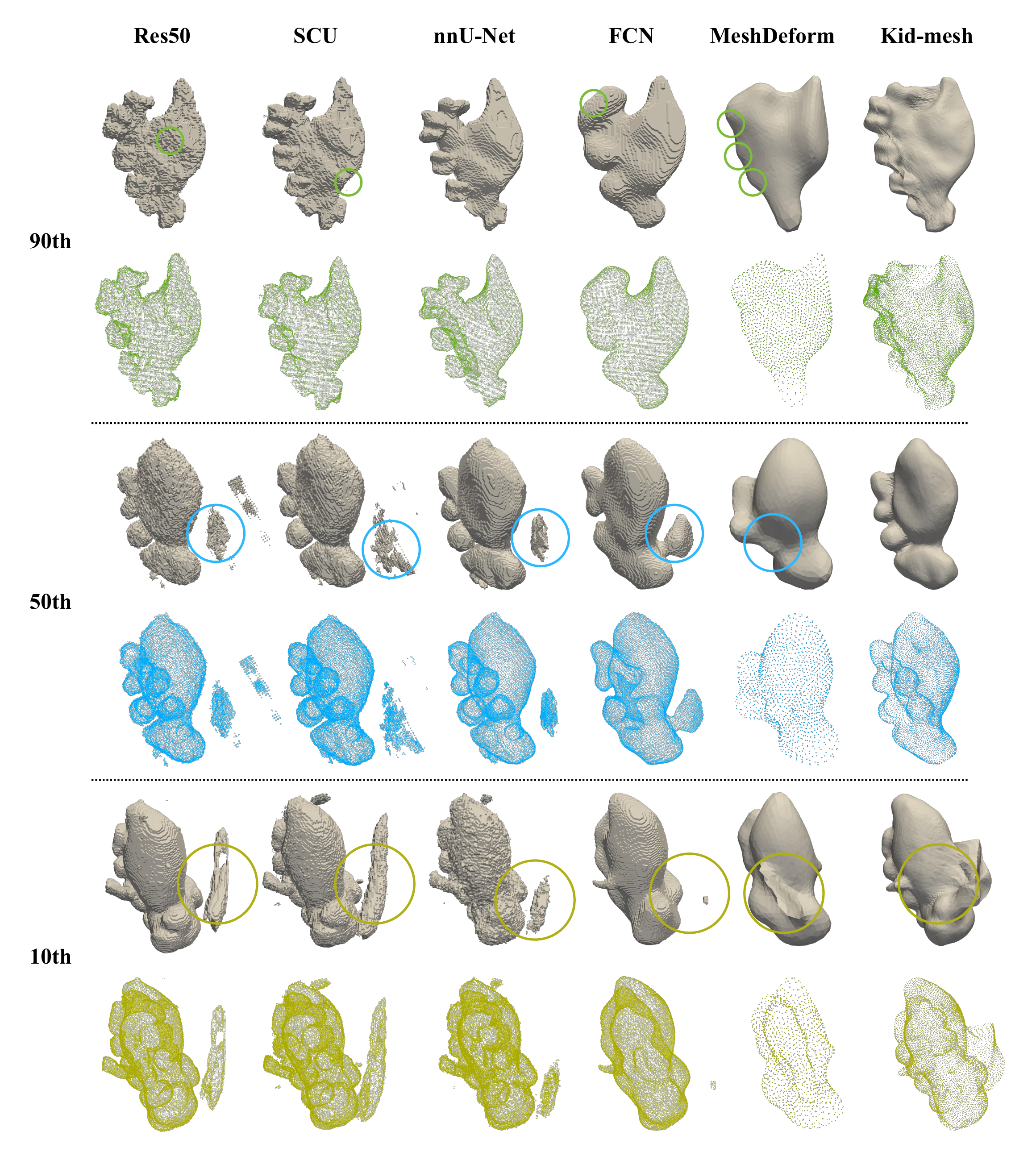}
    \caption{\hly{Visualization of reconstructed meshes from different methods.  
    For each test case, the upper row shows the surface view of meshes, while the lower row illustrates point clouds of meshes.  
    Circles highlight the regions with defects or errors on reconstructed meshes.  
    The label of the baseline has been corrected to “nnU-Net” for consistency.  
    For visual overlay comparisons with pseudo–gold standard surfaces, please refer to Figure~\ref{fig5}, and for quantitative Dice and Jaccard scores, refer to Table~\ref{table1}.}}
    \label{fig4}
\end{figure}

Figure \ref{fig4} displays the 90th, 50th, and 10th percentile cases of different methods in terms of KidMesh’s Dice score. One can see that SCU and RES50 were sensitive to surrounding noise, leading to sharp spikes and outliers on the reconstructed meshes. FCN only captured the general shape of the target region, lacking finer details. Unlike voxel-based reconstruction methods, explicit reconstruction models such as KidMesh produced smoother and clearer meshes, eliminating staircase artifacts. Moreover, although both KidMesh and MeshDeformNet constructed topologically correct smooth meshes, the latter tended to be excessively smooth and less effective for CH mesh reconstruction. 

\begin{figure}[htbp]
    \centering
    \includegraphics[width=0.95\columnwidth]{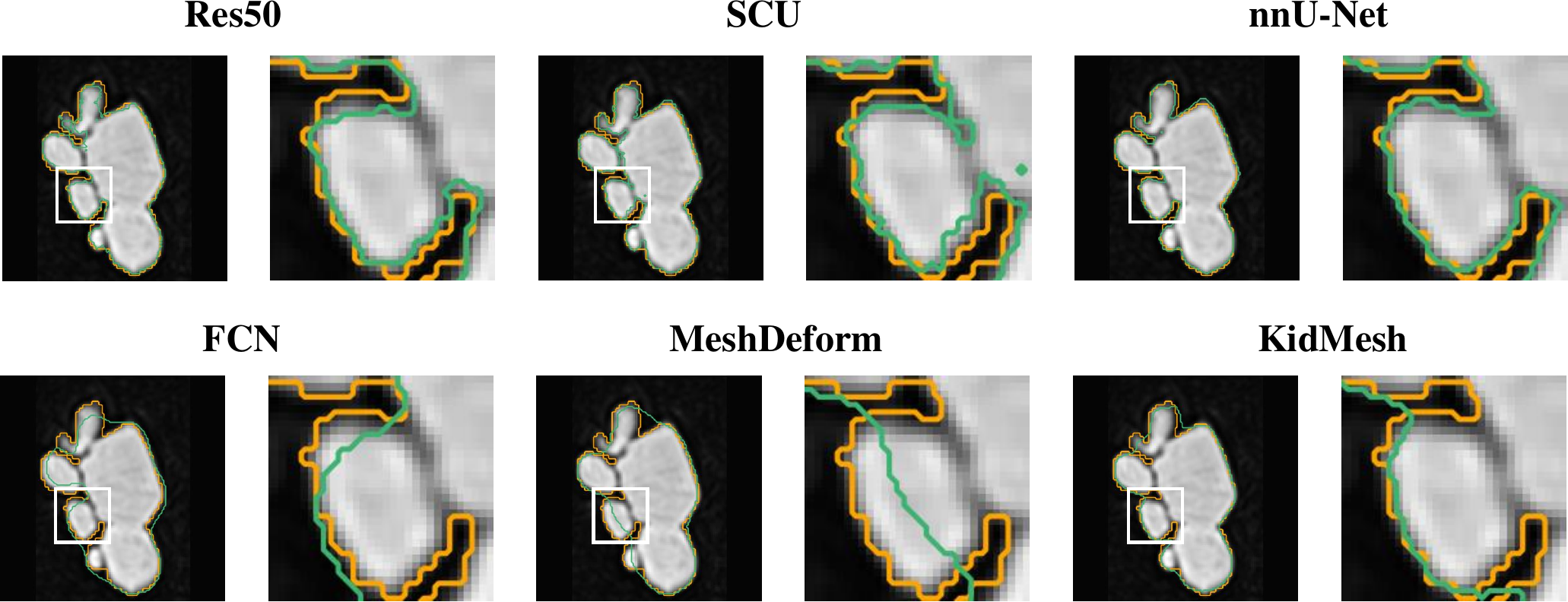}
    \caption{Visualization of segmentation results from different methods. The show results of MeshDeformNet and KidMesh are rasterized after mesh reconstruction. Green lines are the contours of predicted CH regions, while orange lines mark the contours of gold standard CH masks.}
    \label{fig5}
\end{figure}

Figure \ref{fig5} shows the result of voxel-based segmentation methods and rasterized meshes of reconstruction methods. One can observe that voxel-based segmentation methods appeared closer to the gold standard masks. This could be attributed to the fact that they perform segmentation by voxel-wise classification. In contrast, KidMesh employed a deformation-based strategy for mesh reconstruction. The regularization terms (such as Laplacian loss and Edge Length loss) constrained the smoothness of generated meshes, which hindered their ability to accurately capture local geometric details. As a result, it is still challenging to reconstruct regions with large curvature for KidMesh.

\begin{table*}[htbp]
    \centering
    \caption{\hly{Mesh reconstruction results of different methods. The methods above the dash-black line are voxel-based reconstruction methods, while those below the line are explicit reconstruction methods. The time in the first column is the average mesh reconstruction time.}}
    \label{table1}
    \resizebox{\textwidth}{!}{
        \begin{tblr}{
            width = \linewidth,
            colspec = {Q[175]Q[102]Q[115]Q[102]Q[115]Q[110]Q[140]Q[102]},
            cells = {c,m},
            hline{1-2,8} = {-}{},
        }
        \textbf{Method} & \textbf{ASSD$_\text{outer}$↓} & \textbf{ASSD$_\text{mesh}$↓} & \textbf{HD$_\text{outer}$↓} & \textbf{HD$_\text{mesh}$↓} & \textbf{Dice (\%)↑} & \textbf{Jaccard (\%)↑} & \textbf{P2SD↓} \\
        {\textbf{nnU-Net + Post} (71.2s)} & \textbf{1.60±.384} & \textbf{1.41±.640} & 29.51±1.66 & 28.16±1.53 & \textbf{88.3±3.92} & \textbf{81.9±6.63} & 1.98±1.15 \\
        {\textbf{FCN + Post} (61.1s)} & 2.05±.512 & 1.86±.576 & 25.28±2.76 & 24.96±1.21 & 83.0±5.21 & 73.4±8.23 & 1.60±.576 \\
        {\textbf{SCU + Post} (88.3s)} & 2.56±.646 & 2.43±.428 & 37.50±2.87 & 36.48±2.79 & 78.3±4.56 & 67.4±7.23 & 3.52±2.30 \\
        {\textbf{RES50 + Post} (86.5s)} & 1.79±.448 & 1.60±.640 & 31.80±1.77 & 31.36±1.54 & 87.5±3.56 & 80.5±6.03 & 2.18±1.02 \\ \hline[dashed]
        {\textbf{MeshDeformNet} (3.14s)} & 2.30±.640 & 2.18±.512 & 10.42±.640 & 9.31±.576 & 84.3±3.14 & 75.3±3.95 & 1.54±.768 \\
        {\textbf{KidMesh} (0.36s)} & 2.11±.256 & 2.04±.320 & \textbf{8.22±.576} & \textbf{8.16±.640} & 86.0±3.58 & 78.3±2.09 & \textbf{1.15±.192} \\
        \end{tblr}
    }
\end{table*}

\subsection{\hlb{Surface Validity Study}}
\label{sec:Topology Correctness}

\begin{figure}[htbp]
    \centering
    \includegraphics[width=0.95\columnwidth]{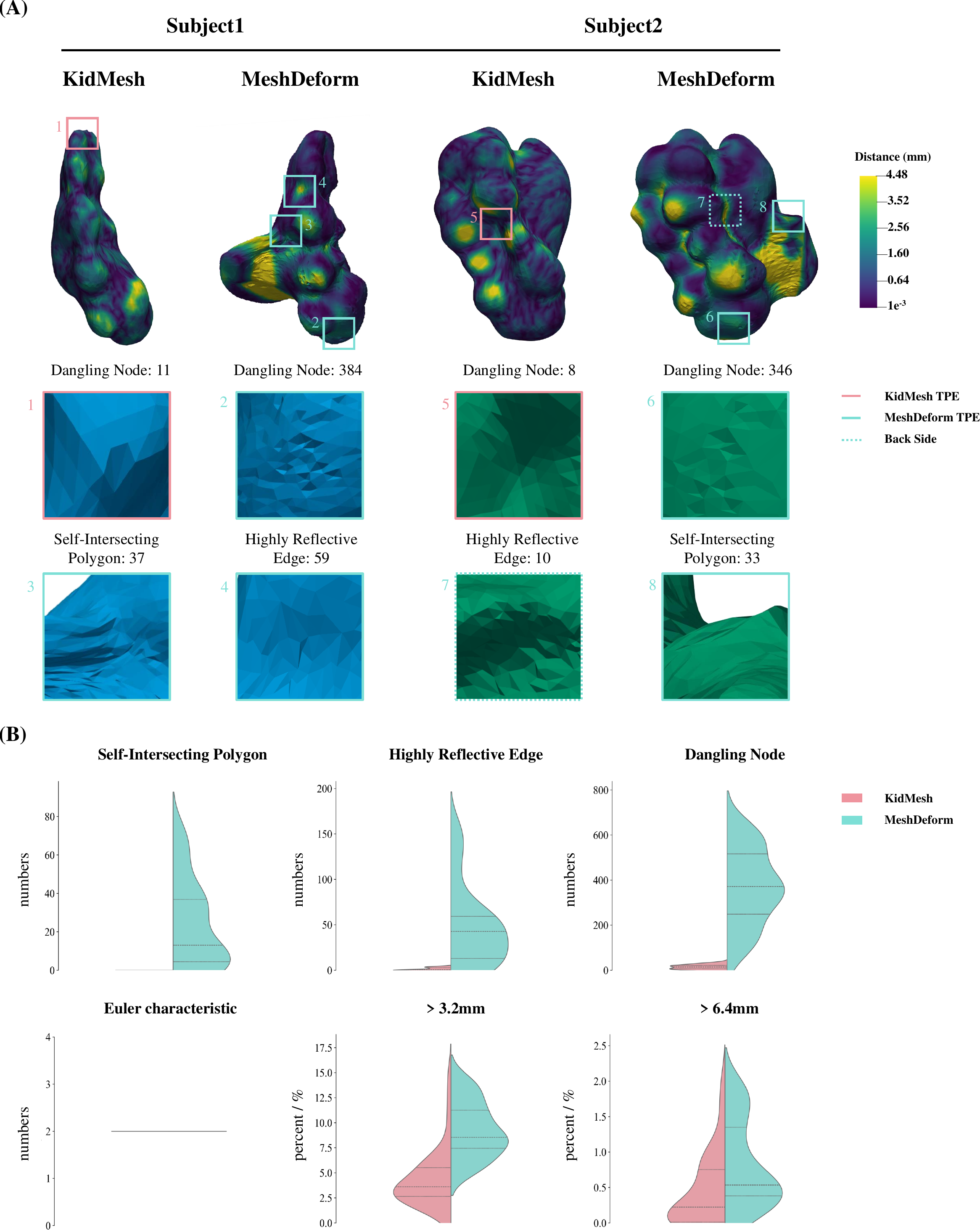}
    \caption{\hlb{Comparison of KidMesh and MeshDeformNet in terms of surface validity. 
    (A) The colored models visualize the Euclidean distance between the predicted meshes and the pseudo--gold standard meshes. The lower panel illustrates three representative surface validity failures observed in the reconstructed meshes, including dangling nodes and highly reflective edges, which correspond to geometric degeneracies, as well as self-intersecting polygons, which represent genuine topological invalidities. 
    (B) Statistical comparison of KidMesh and MeshDeformNet with respect to surface validity failures, Euler characteristics, and the proportion of vertices whose distances exceed 3.2~mm and 6.4~mm.}}
    \label{fig7}
\end{figure}

\hlb{We further evaluated surface validity failures related to topological correctness in meshes reconstructed by KidMesh, including dangling nodes~\cite{attene2010lightweight}, self-intersecting polygons~\cite{botsch2010polygon}, and highly reflective edges. Although mesh connectivity is preserved throughout deformation and upsampling, geometric deformation may still compromise surface validity and lead to failures that affect manifoldness or watertightness in practice. In this study, dangling nodes and highly reflective edges correspond to geometric degeneracies, including degenerate local neighborhoods and severe local geometric distortion, whereas self-intersecting polygons represent genuine topological invalidities that arise from geometric deformation despite unchanged connectivity. Such surface validity failures can introduce geometric instability and numerical artifacts in downstream fluid simulations.} Figure~\ref{fig7}(A) presents representative cases of the three failure types, where dangling nodes are mainly concentrated near protruding boundary regions and self-intersecting polygons and highly reflective edges tend to occur in areas with pronounced curvature changes. In Figure~\ref{fig7}(B), the distance denotes the Euclidean distance from vertices associated with surface validity failures to the nearest points on the pseudo-gold standard mesh surface, quantifying the geometric deviation of invalid regions from the anatomical boundary. Although KidMesh and MeshDeformNet yield identical Euler characteristics~\cite{leinster2008euler,gopinath2023learning}, KidMesh exhibits clear advantages in reducing other surface validity failures and shows a lower proportion of vertices with deviations exceeding 3.2~mm and 6.4~mm, demonstrating improved geometric stability and topological correctness under deformation compared with the state-of-the-art reconstruction method MeshDeformNet.

\subsection{Ablation Studies}
\label{sec:Ablation}
\hly{
To assess the contribution of each component within the proposed KidMesh framework, 
we conducted ablation experiments using the same training and evaluation settings as those applied in the main study. 
Specifically, we examined three aspects: 
(1) the effect of the learned neighborhood sampling and self-attention fusion in FSM, 
(2) the influence of the coarse-to-fine deformation strategy, 
and (3) the contribution of each loss term in the reconstruction loss \(L_{\mathrm{Rec}}\), 
including Chamfer, Normal, Laplacian, Edge, Area, and Seal losses. 
Each variant was trained independently under identical hyperparameters and evaluated on the same validation and test split 
using both mesh- and surface-level metrics (ASSD\(_\text{mesh}\), HD\(_\text{mesh}\), Dice, Jaccard, and P2SD).
}

\begin{table*}[htbp]
    \centering
    \caption{\hly{Ablation study on key components of KidMesh. Lower is better for ASSD/HD/P2SD, higher is better for Dice/Jaccard.}}
    \label{tab:ablation}
    \resizebox{\textwidth}{!}{
        \begin{tblr}{
                width = \linewidth,
                colspec = {Q[195]Q[115]Q[115]Q[115]Q[115]Q[115]Q[115]Q[115]},
                cells = {c,m},
                hline{1-2,12} = {-}{},
            }
            \textbf{Variant} & \textbf{ASSD$_\text{outer}$↓} & \textbf{ASSD$_\text{mesh}$↓} & \textbf{HD$_\text{outer}$↓} & \textbf{HD$_\text{mesh}$↓} & \textbf{Dice (\%)↑} & \textbf{Jaccard (\%)↑} & \textbf{P2SD↓} \\
            {\textbf{KidMesh}} & \textbf{2.11±.256} & \textbf{2.04±.320} & \textbf{8.22±.576} & \textbf{8.16±.640} & \textbf{86.0±3.58} & \textbf{78.3±2.09} & \textbf{1.15±.192} \\ \hline[dashed]
            {\textbf{w/o learned neighborhood}} & 2.46±.311 & 2.37±.398 & 9.38±.644 & 9.15±.526 & 83.9±3.21 & 75.6±3.82 & 1.42±.270 \\
            {\textbf{w/o self-attention (in FSM)}} & 2.33±.295 & 2.25±.356 & 8.96±.612 & 8.75±.587 & 84.7±3.16 & 76.3±3.90 & 1.31±.228 \\
            {\textbf{single-stage deformation}} & 2.52±.327 & 2.46±.374 & 9.70±.662 & 9.41±.611 & 83.2±3.45 & 74.9±4.11 & 1.48±.255 \\ \hline[dashed]
            {\textbf{w/o Chamfer loss}} & 2.59±.338 & 2.48±.392 & 9.92±.688 & 9.53±.623 & 82.6±3.59 & 74.2±4.35 & 1.52±.286 \\
            {\textbf{w/o Normal loss}} & 2.44±.320 & 2.35±.361 & 9.24±.630 & 9.02±.601 & 84.1±3.33 & 75.8±3.92 & 1.36±.241 \\
            {\textbf{w/o Laplacian loss}} & 2.28±.299 & 2.19±.340 & 8.84±.604 & 8.66±.580 & 85.3±3.27 & 77.1±3.65 & 1.23±.210 \\
            {\textbf{w/o Edge length loss}} & 2.32±.303 & 2.23±.348 & 8.90±.615 & 8.70±.592 & 84.9±3.35 & 76.8±3.73 & 1.26±.219 \\
            {\textbf{w/o Area loss}} & 2.25±.286 & 2.17±.333 & 8.79±.593 & 8.60±.569 & 85.6±3.24 & 77.3±3.58 & 1.21±.207 \\
            {\textbf{w/o Seal loss}} & 2.30±.298 & 2.22±.342 & 8.83±.606 & 8.65±.582 & 85.1±3.29 & 76.9±3.66 & 1.25±.213 \\
        \end{tblr}
    }
\end{table*}

\hly{
As shown in Table~\ref{tab:ablation}, removing either the learned neighborhood sampling or the self-attention module in FSM led to a clear decline in performance, demonstrating that both local and global vertex aggregation are vital for accurate feature propagation. 
The self-attention mechanism, in particular, enhances long-range geometric consistency by enabling information exchange between spatially distant but anatomically} \hly{related vertices. 
Disabling this component reduced the Dice score by 1.3\% and increased P2SD by 0.16~mm, while introducing only minor computational overhead (an additional 7\% GPU memory usage and 0.03~s in inference time per case). 
These findings confirm that the parameter-free attention in FSM effectively reinforces global coherence with minimal computational cost.} 

\hly{Removing the coarse-to-fine deformation strategy resulted in less accurate boundary reconstruction, indicating that progressive refinement is essential for capturing precise geometric structures. 
Regarding the reconstruction loss, excluding the Chamfer or Normal terms caused a significant increase in distance-based errors (ASSD and HD), whereas the Laplacian, Edge, Area, and Seal losses mainly contributed to smoother surfaces and greater topological stability. 
Taken together, these results demonstrate that every component plays a meaningful role in enhancing the robustness and accuracy of KidMesh, and that the self-attention mechanism within FSM is especially important for maintaining topological continuity and producing CFD-ready meshes.
}

\subsection{Urodynamic Simulation Study}
\label{sec:Fluid Simulation}
We utilized CFD techniques to simulate urine flow velocity and pressure based on reconstructed CH meshes. Material properties and simulation model parameters were configured based on previous research studies \cite{inman2013impact,jang2022numerical,maralescu2022non}. Specifically, the properties of urine and surrounding tissues are listed in Table \ref{table4}. Meanwhile, energy model and k-epsilon turbulence model were employed to simulate the fluid dynamics. The final simulation results were extracted after 200 iterations. \hlb{Moreover, a constant velocity boundary condition of 0.0001 m/s was prescribed at the inlet, a zero-gauge pressure condition was applied at the outlet, and no-slip boundary conditions were imposed on all vessel walls \cite{yang2025non}.}

\begin{figure}[htbp]
    \centering
    \includegraphics[width=0.9\columnwidth]{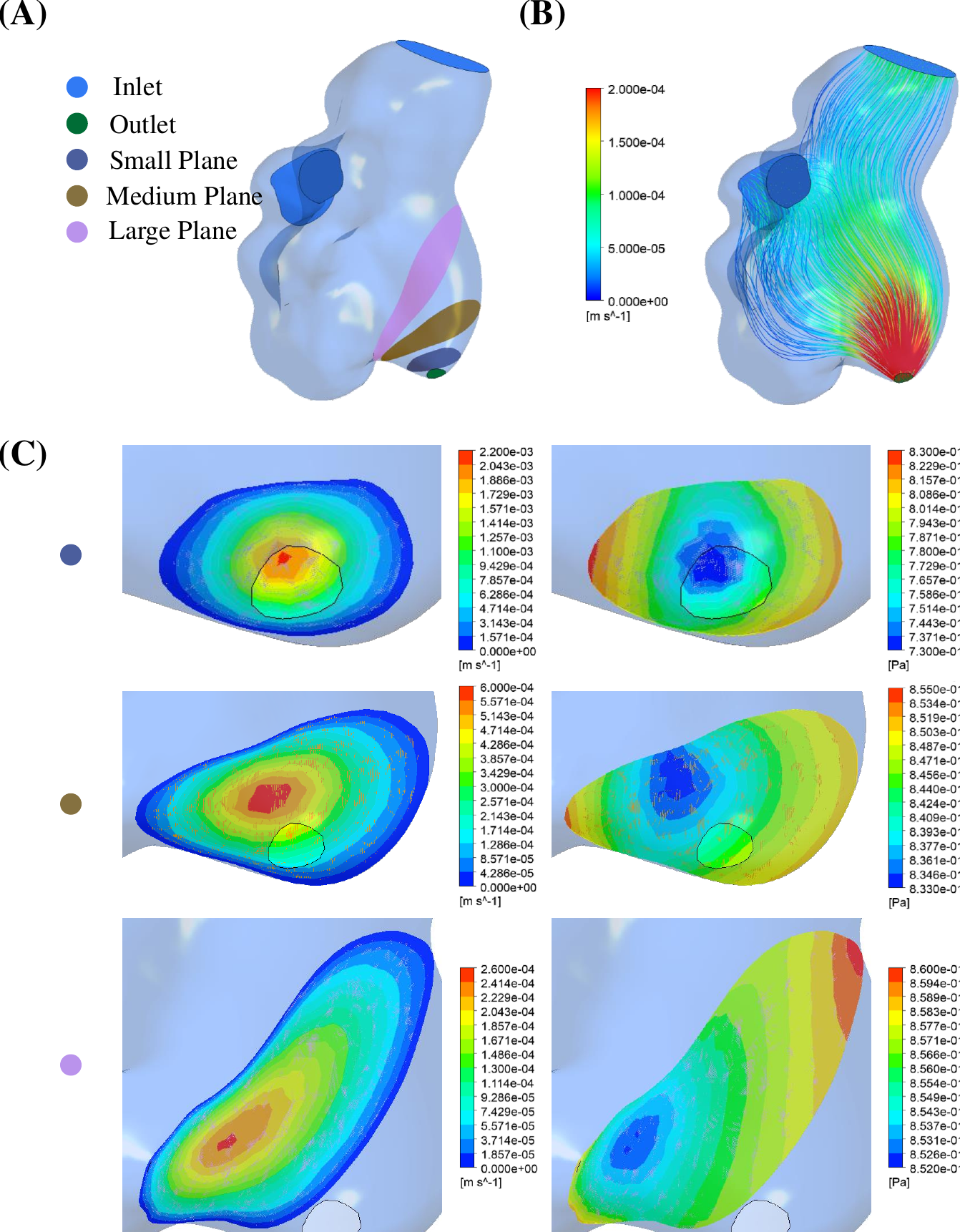}
    \caption{\hly{The simulated urine flow velocity and pressure on the reconstructed CH meshes. 
    (A) The geometry of the reconstructed renal pelvis mesh for a CH patient, which defines the CFD simulation domain, with the inlet at the renal pelvis opening and the outlet at the ureteropelvic junction.  Configurations of the CFD simulation were specified in Table \ref{table4}
    Regions labeled as small (dark blue), medium (brown), and large (purple) indicate different planar sections for analysis.} 
    (B) The flow velocity streamlines in the renal model, showing dynamic urine flow from the inlet (blue) to the outlet (green). 
    (C) The velocity (left column) and pressure (right column) maps of the small, medium, and large planes.}
    \label{fig8}
\end{figure}

\hlb{The KidMesh-reconstructed meshes were directly used for CFD simulations without any additional mesh repair or post-processing.} Figure \ref{fig8} (A) shows a CH mesh reconstructed by KidMesh, which was adopted for CFD simulation. The simulation produced plausible urine flow results, as displayed in Figure \ref{fig8} (B). Additionally, three planes of varying sizes (small, medium, and large) were extracted for visualization, as depicted in Figure \ref{fig8} (C). One can observe that the velocity and pressure were reasonable, as they exhibited close relationships to the morphology of the mesh. This indicated that KidMesh effectively captured the morphology renal pelvis, demonstrating its capability to produce computational meshes for CFD simulations.

\hly{
To further validate the fidelity of the reconstructed meshes, we conducted comparative CFD simulations using both KidMesh-generated meshes and pseudo–gold standard meshes derived from manual segmentation. 
The resulting velocity and pressure fields showed high quantitative agreement between the two mesh types. 
Across the entire renal pelvis lumen, the mean velocity magnitude differed by less than 3.6\%, and the average pressure discrepancy was below 1.0~Pa. 
Both simulations exhibited similar streamline topologies, including flow acceleration at the ureteropelvic junction and pressure gradients along the central lumen, confirming that KidMesh accurately preserves the hydraulic resistance and energy dissipation characteristics of the urinary tract. 
These results validate that KidMesh meshes can reliably reproduce physiological urodynamics comparable to those computed from manual reference geometries, supporting their suitability for CFD-based analysis without additional manual correction.
}

\begin{figure}[htbp]
    \centering
    \includegraphics[width=0.8\columnwidth]{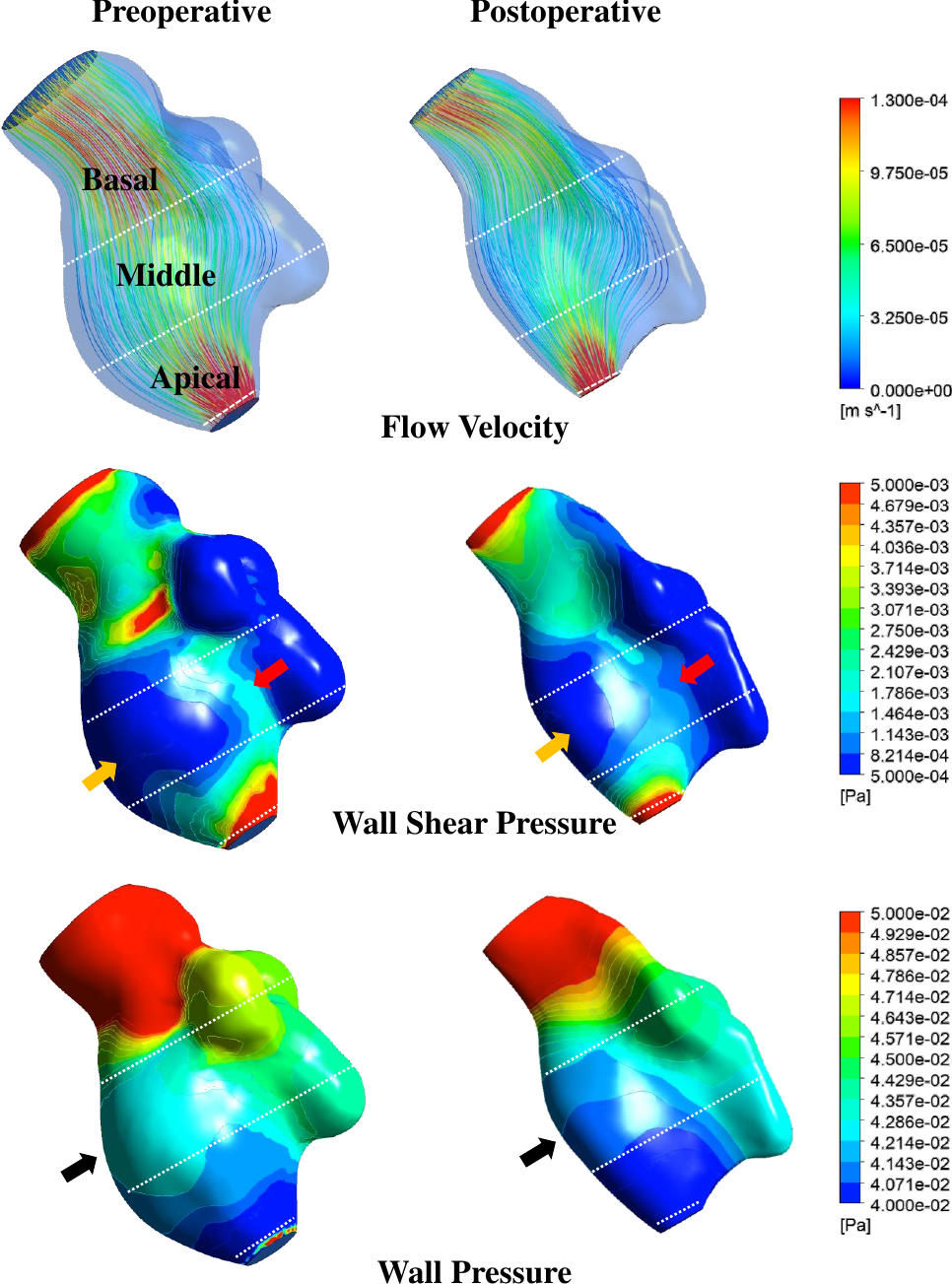}
    \caption{Comparison of urodynamic characteristics of preoperative and postoperative CH meshes. This figure shows the results of urine flow velocity (first row), wall shear pressure (second row), and wall pressure (third row).}
    \label{fig9}
\end{figure}

We further conducted simulations to explore the changes in urodynamic information for the same patient. As shown in Figure \ref{fig9}, both preoperative and postoperative CH meshes may exhibit slower or faster in terms of flow velocity in the middle (wide) and apical (narrow) regions. Thus, it is difficult to evaluate the effectiveness of surgery from urine flow velocity. Regarding wall shear stress, one can observe that the stress values in the areas marked by the red and orange arrows became more similar after surgery. This indicated that the surgery improved wall shear stress. While this suggested that wall shear stress could be used as an index to assess the effectiveness of the surgery, its significance is limited. Furthermore, one can see the pressure was largely reduced in the middle areas of the renal pelvis after surgery, as indicated by the black arrows. This was consistent with the surgical relief. Overall, by simulation with preoperative and postoperative meshes, we could provide urodynamic evidence for the efficacy of surgical intervention.  This may offer a new technique to assess surgical outcomes in clinical practice.

\hly{The material parameters used for CFD simulations, such as urine density and viscosity, were chosen within physiologically reported ranges through multiple trials to ensure numerical stability and convergence.  
The final values adopted for all experiments are summarized in Table~\ref{table4}.  
To assess robustness against physiological variability, we further perturbed both parameters by \(\pm10\%\) and reran the simulations.  
The resulting velocity and pressure distributions exhibited less than 3.5\% deviation on average, and the streamline topology remained qualitatively consistent.  
These results indicate that the CFD pipeline based on KidMesh-reconstructed meshes is both numerically stable and robust for urodynamic evaluation.}

\begin{table}[hbp]
    \centering
    \caption{The configuration of CFD simulation experiment on the generated meshes of KidMesh.}
    \label{table4}
    \resizebox{0.9\columnwidth}{!}{
        \begin{tblr}{
                width = \linewidth,
                colspec = {Q[l,40]Q[c,30]Q[c,30]},
                hline{1-2,7} = {-}{},
            }
            \textbf{Parameters}                        & \textbf{Urine} & \textbf{Tissue} \\
            \textbf{Density (kg/m\textsuperscript{3})} & 1050           & 1050            \\
            \textbf{Specific Heat (J/(kg·K))}          & 4180           & 3800            \\
            \textbf{Thermal Conductivity (W/(m·K))}    & 0.6            & 0.5             \\
            \textbf{Viscosity (kg/(m·s))}              & 0.002          &  -              \\
            \textbf{Initial Velocity (m/s)}            & 0.0001         &  -              \\
        \end{tblr}
    }
\end{table}


\section{Discussion}
\label{sec:Discussion}
KidMesh presents an end-to-end explicit mesh reconstruction framework that directly generates CFD-ready renal meshes from MRU images.  
By integrating the FEM, FSM, and MDM, it bridges the gap between volumetric medical imaging and simulation-ready mesh representation.  
This capability brings tangible clinical value, especially for organs such as the kidneys and heart, where fluid simulations provide quantitative insights into postoperative outcomes and flow dynamics~\cite{mascolini2023influence,africa2024lifex}.  
Our experimental evaluations demonstrate that KidMesh achieves high geometric fidelity and computational efficiency, outperforming recent explicit reconstruction frameworks in both accuracy and topological correctness.  
Furthermore, CFD simulations performed on reconstructed renal geometries confirmed that the generated meshes preserve flow characteristics comparable to those derived from expert-refined pseudo–gold standard meshes, thereby validating their practical utility for functional urodynamic analysis.


\hly{While our work is conceptually related to pioneering template-based methods for cortical surface reconstruction (e.g., Vox2Cortex~\cite{Bongratz2022}) and whole-heart modeling (e.g., MeshDeformNet~\cite{Kong2021WholeHeart}), KidMesh is differentiated by several key design choices tailored for CH, as summarised in Table~\ref{tab:mesh_based_comparison}. First, it employs a multi-stage, hierarchical deformation strategy, which integrates a dynamic VF module, to handle the highly heterogeneous geometry of the renal pelvis and calyces. This coarse-to-fine approach, which adaptively increases mesh resolution, has not been adequately addressed in the literature. Second, it operates under weak supervision, learning from pseudo-gold standard meshes derived from voxel masks, which circumvents the need for impractical mesh-level annotations. Finally, it incorporates explicit topology-preserving losses to generate watertight, manifold meshes that are directly suitable for downstream urodynamic CFD simulations. These innovations collectively address the specific anatomical and functional requirements of CH analysis.}

\begin{table*}[!t]
\renewcommand{\arraystretch}{1.2}
\centering
\caption{\hly{Comparison of recent mesh-based surface reconstruction frameworks in medical imaging.}}
\label{tab:mesh_based_comparison}
\resizebox{\linewidth}{!}{
\begin{tabular}{lcccccc}
\toprule
\textbf{Method} & \textbf{Targets} &  \textbf{Publication} & \textbf{Supervision (Mesh GT)} & \textbf{Topology \& Watertightness} & \textbf{Code} & \textbf{Notes} \\
\midrule
Vox2Cortex~\cite{Bongratz2022} & Cortex & CVPR  & Yes & Topology-correct, fast & Yes & Neuro MRI, relies on FreeSurfer \\
V2C-Flow~\cite{Bongratz2024} & Cortex &  MedIA & Yes & Joint WM/pial, consistent deformation & — & Continuous deformation fields \\
PialNN~\cite{Ma2021} & Cortex & MICCAI & Yes & High-res., topology preserved & Yes & Deform WM $\rightarrow$ pial surface \\
Voxel2Mesh~\cite{Wickramasinghe2020} & Liver \& Hippocampus  & MICCAI & Yes & End-to-end, moderate control & Yes & Direct volume-to-mesh conversion \\
MeshDeformNet~\cite{kong2021deep} & Heart & IPMI & Yes & Smooth, simulation-suitable & Yes & Organ-level template deformation \\
HeartDeformNets~\cite{kong2022learning} & Heart & TMI & Yes & Watertight, smooth surfaces & Yes & Template deformation via biharmonic coords \\
UNetFlow~\cite{Bongratz2023} & Abdomen & Sci. Rep. & Yes & Diffeomorphic, topology-preserving & — & Homeomorphic deformation constraint \\
\midrule
\textbf{KidMesh (ours)} & Renal Pelvis and Calyces &  & No & Watertight, topology-regularized & Yes & CFD-ready, curvature-adaptive design \\
\bottomrule
\end{tabular}}
\\[0.3em]
\begin{minipage}{0.8\linewidth}
\footnotesize \textit{Note.} “Mesh GT” indicates whether high-quality mesh annotations are required for supervision. 
\end{minipage}
\end{table*}

Despite these advantages, several limitations still need to be addressed.  
First, the model shows reduced accuracy in regions with high curvature or fine anatomical structures, such as the narrow renal calyces. 
\hlb{Future improvements may incorporate curvature-aware or normal-consistency losses~\cite{vetsch2022neuralmeshing,soboleva2023prs,hernandez2025sr,zhang2025high} during the fine-deformation stage to better preserve thin, high-curvature structures, together with adaptive mesh refinement strategies~\cite{fu2024lfs} that allocate higher vertex density to geometrically complex regions, while keeping topology-aware regularization as the dominant constraint.}
Second, although the topology of reconstructed meshes is generally well maintained, minor artifacts such as discontinuous protrusions or dangling nodes can still appear (see Section~\ref{sec:Topology Correctness}).  
Previous studies have indicated that differential homeomorphism constraints can effectively reduce these errors~\cite{Bongratz2023}. 
Incorporating such constraints into the KidMesh framework may further enhance reconstruction robustness.  
Third, the current study is based on a non-public single-center MRU dataset, which limits the model's generalization to data acquired from different scanners or imaging protocols.  
Future work will focus on building a multi-center MRU dataset to improve cross-domain robustness and support external validation.  


\section{Conclusion}
\label{sec:Conclusion}
\hly{In summary, KidMesh introduces an end-to-end explicit deep-learning framework for generating CFD-ready renal meshes directly from MRU data. By combining hierarchical coarse-to-fine deformation, a weakly supervised training strategy, and topology-preserving regularization, the model achieves accurate geometric reconstruction and watertight surface topology that support reliable urodynamic analysis.
Although challenges remain, including the need to improve performance in high-curvature regions and enhance generalization across imaging centers, this work lays a strong foundation for future research. A promising future direction is to extend the KidMesh framework to model the entire urinary tract, including both upper (e.g., ureters) and lower (e.g., bladder) urinary system structures. This would enable comprehensive, patient-specific simulations of urinary dynamics, further advancing clinically applicable and simulation-oriented mesh reconstruction.}

\section*{Ethics Statement}
This study was carried out in compliance with applicable laws and institutional guidelines and received approval from the Ethics Committee of Fujian Provincial Hospital (Approval No. K2022-04-012, approved on April 12, 2022). All procedures involving human participants followed ethical standards, and written informed consent was obtained from all participants or their legal guardians prior to their inclusion in the study. The privacy rights of all participants were rigorously safeguarded. No \hlb{public} datasets were used in this study. There are no conflicts of interest related to ethical considerations in this study.

\section*{References}

\bibliographystyle{IEEEtran}
\bibliography{ref1.bib}

@article{kong2021deep,
  title={A deep-learning approach for direct whole-heart mesh reconstruction},
  author={Kong, Fanwei and Wilson, Nathan and Shadden, Shawn},
  journal={Medical image analysis},
  volume={74},
  pages={102222},
  year={2021},
  publisher={Elsevier}
}

@article{africa2024lifex,
  title={lifex-cfd: An open-source computational fluid dynamics solver for cardiovascular applications},
  author={Africa, Pasquale Claudio and Fumagalli, Ivan and Bucelli, Michele and Zingaro, Alberto and Fedele, Marco and Quarteroni, Alfio and others},
  journal={Computer Physics Communications},
  volume={296},
  pages={109039},
  year={2024},
  publisher={Elsevier}
}

@article{mascolini2023influence,
  title={Influence of transurethral catheters on urine pressure-flow relationships in males: A computational fluid-dynamics study},
  author={Mascolini, Maria Vittoria and Fontanella, Chiara Giulia and Berardo, Alice and Carniel, Emanuele Luigi},
  journal={Computer Methods and Programs in Biomedicine},
  volume={238},
  pages={107594},
  year={2023},
  publisher={Elsevier}
}

@article{Adams1994,
  title = {Seeded Region Growing},
  author = {Adams, R. and Bischof, L.},
  year = {1994},
  month = jun,
  journal = {IEEE Transactions on Pattern Analysis and Machine Intelligence},
  volume = {16},
  number = {6},
  pages = {641--647},
  doi = {10.1109/34.295913},
  lccn = {1}
}

@article{onen2020grading,
	title={Grading of hydronephrosis: an ongoing challenge},
	author={Onen, Abdurrahman},
	journal={Frontiers in pediatrics},
	volume={8},
	pages={538943},
	year={2020},
	publisher={Frontiers}
}

@article{kohno2020pediatric,
	title={Pediatric congenital hydronephrosis (ureteropelvic junction obstruction): Medical management guide},
	author={Kohno, Miyuki and Ogawa, Tetsushi and Kojima, Yoshiyuki and Sakoda, Akiko and Johnin, Kazuyoshi and Sugita, Yoshifumi and Nakane, Akihiro and Noguchi, Mitsuru and Moriya, Kimihiko and Hattori, Motoshi and others},
	journal={International Journal of Urology},
	volume={27},
	number={5},
	pages={369--376},
	year={2020},
	publisher={Wiley Online Library}
}

@article{gopinath2023learning,
  title={Learning joint surface reconstruction and segmentation, from brain images to cortical surface parcellation},
  author={Gopinath, Karthik and Desrosiers, Christian and Lombaert, Herve},
  journal={Medical Image Analysis},
  volume={90},
  pages={102974},
  year={2023},
  publisher={Elsevier}
}

@article{Balin2023,
  title = {Layer-Neighbor Sampling --- Defusing Neighborhood Explosion in GNNs},
  author = {Balin, Muhammed Fatih and {\c C}ataly{\"u}rek, {\"U}mit},
  year = {2023},
  month = dec,
  journal = {Advances in Neural Information Processing Systems},
  volume = {36},
  pages = {25819--25836}
}

@inproceedings{Bongratz2022,
  title = {Vox2Cortex: Fast Explicit Reconstruction of Cortical Surfaces from 3D MRI Scans with Geometric Deep Neural Networks},
  shorttitle = {Vox2Cortex},
  booktitle = {2022 IEEE/CVF Conference on Computer Vision and Pattern Recognition (CVPR)},
  author = {Bongratz, Fabian and Rickmann, Anne-Marie and Polsterl, Sebastian and Wachinger, Christian},
  year = {2022},
  month = jun,
  pages = {20741--20751},
  publisher = {IEEE},
  address = {New Orleans, LA, USA},
  doi = {10.1109/CVPR52688.2022.02011},
  copyright = {https://doi.org/10.15223/policy-029},
  isbn = {978-1-66546-946-3}
}

@article{Bongratz2023,
  title = {Abdominal Organ Segmentation via Deep Diffeomorphic Mesh Deformations},
  author = {Bongratz, Fabian and Rickmann, Anne-Marie and Wachinger, Christian},
  year = {2023},
  month = oct,
  journal = {Scientific Reports},
  volume = {13},
  number = {1},
  pages = {18270},
  doi = {10.1038/s41598-023-45435-2},
  lccn = {3}
}

@article{Bongratz2024,
  title = {Neural Deformation Fields for Template-Based Reconstruction of Cortical Surfaces from MRI},
  author = {Bongratz, Fabian and Rickmann, Anne-Marie and Wachinger, Christian},
  year = {2024},
  month = apr,
  journal = {Medical Image Analysis},
  volume = {93},
  pages = {103093},
  doi = {10.1016/j.media.2024.103093},
  lccn = {1}
}

@article{Cheadle2003,
  title = {Analysis of Microarray Data Using Z Score Transformation},
  author = {Cheadle, Chris and Vawter, Marquis P. and Freed, William J. and Becker, Kevin G.},
  year = {2003},
  month = may,
  journal = {The Journal of Molecular Diagnostics},
  volume = {5},
  number = {2},
  pages = {73--81},
  doi = {10.1016/S1525-1578(10)60455-2},
  copyright = {https://www.elsevier.com/tdm/userlicense/1.0/},
  lccn = {3}
}

@misc{Chen2019,
  title = {Med3D: Transfer Learning for 3D Medical Image Analysis},
  shorttitle = {Med3D},
  author = {Chen, Sihong and Ma, Kai and Zheng, Yefeng},
  year = {2019},
  month = apr,
  journal = {arXiv e-prints},
  doi = {10.48550/arXiv.1904.00625}
}

@inproceedings{Cruz2021,
  title = {DeepCSR: A 3D Deep Learning Approach for Cortical Surface Reconstruction},
  shorttitle = {DeepCSR},
  booktitle = {2021 IEEE Winter Conference on Applications of Computer Vision (WACV)},
  author = {Cruz, Rodrigo Santa and Lebrat, Leo and Bourgeat, Pierrick and Fookes, Clinton and Fripp, Jurgen and Salvado, Olivier},
  year = {2021},
  month = jan,
  pages = {806--815},
  doi = {10.1109/WACV48630.2021.00085}
}

@article{Fischl2012a,
  title = {FreeSurfer},
  author = {Fischl, Bruce},
  year = {2012},
  month = aug,
  journal = {NeuroImage},
  series = {20 YEARS OF fMRI},
  volume = {62},
  number = {2},
  pages = {774--781},
  doi = {10.1016/j.neuroimage.2012.01.021},
  lccn = {1}
}

@article{Garcia-Valtuille2006,
  title = {Magnetic Resonance Urography: A Pictorial Overview},
  shorttitle = {Magnetic Resonance Urography},
  author = {{Garc{\'i}a-Valtuille}, R and {Garc{\'i}a-Valtuille}, A I and Abascal, F and Cerezal, L and Arg{\"u}ello, M C},
  year = {2006},
  month = jul,
  journal = {The British Journal of Radiology},
  volume = {79},
  number = {943},
  pages = {614--626},
  publisher = {The British Institute of Radiology},
  doi = {10.1259/bjr/21075982},
  lccn = {3}
}

@inproceedings{Genova2019,
  title = {Learning Shape Templates With Structured Implicit Functions},
  booktitle = {Proceedings of the IEEE/CVF International Conference on Computer Vision},
  author = {Genova, Kyle and Cole, Forrester and Vlasic, Daniel and Sarna, Aaron and Freeman, William T. and Funkhouser, Thomas},
  year = {2019},
  pages = {7154--7164}
}

@inproceedings{Gohil2021,
  title = {Kidney and Kidney Tumor Segmentation Using Spatial and Channel Attention Enhanced U-Net},
  booktitle = {Kidney and Kidney Tumor Segmentation: MICCAI 2021 Challenge, KiTS 2021, Held in Conjunction with MICCAI 2021, Strasbourg, France, September 27, 2021, Proceedings},
  author = {Gohil, Sajan and Lad, Abhi},
  year = {2021},
  month = sep,
  pages = {151--157},
  publisher = {Springer-Verlag},
  address = {Berlin, Heidelberg},
  doi = {10.1007/978-3-030-98385-7_20},
  isbn = {978-3-030-98384-0}
}

@article{Guillard2024,
  title = {DeepMesh: Differentiable Iso-Surface Extraction},
  shorttitle = {DeepMesh},
  author = {Guillard, Beno{\^i}t and Remelli, Edoardo and Lukoianov, Artem and Yvernay, Pierre and Richter, Stephan R. and Bagautdinov, Timur and Baque, Pierre and Fua, Pascal},
  year = {2024},
  journal = {IEEE Transactions on Pattern Analysis and Machine Intelligence},
  pages = {1--15},
  doi = {10.1109/TPAMI.2024.3392291},
  lccn = {1}
}

@article{Henschel2020,
  title = {FastSurfer - A Fast and Accurate Deep Learning Based Neuroimaging Pipeline},
  author = {Henschel, Leonie and Conjeti, Sailesh and Estrada, Santiago and Diers, Kersten and Fischl, Bruce and Reuter, Martin},
  year = {2020},
  month = oct,
  journal = {NeuroImage},
  volume = {219},
  pages = {117012},
  doi = {10.1016/j.neuroimage.2020.117012},
  lccn = {1}
}

@article{chen2024neural,
  title={Neural implicit surface reconstruction of freehand 3D ultrasound volume with geometric constraints},
  author={Chen, Hongbo and Kumaralingam, Logiraj and Zhang, Shuhang and Song, Sheng and Zhang, Fayi and Zhang, Haibin and Pham, Thanh-Tu and Punithakumar, Kumaradevan and Lou, Edmond HM and Zhang, Yuyao and others},
  journal={Medical Image Analysis},
  volume={98},
  pages={103305},
  year={2024},
  publisher={Elsevier}
}

@inproceedings{Hu2021,
  title = {Self-Supervised 3D Mesh Reconstruction From Single Images},
  booktitle = {Proceedings of the IEEE/CVF Conference on Computer Vision and Pattern Recognition},
  author = {Hu, Tao and Wang, Liwei and Xu, Xiaogang and Liu, Shu and Jia, Jiaya},
  year = {2021},
  pages = {6002--6011}
}

@article{Huo2021,
  title = {Segmentation of Whole Breast and Fibroglandular Tissue Using nnU-Net in Dynamic Contrast Enhanced MR Images},
  author = {Huo, Lu and Hu, Xiaoxin and Xiao, Qin and Gu, Yajia and Chu, Xu and Jiang, Luan},
  year = {2021},
  month = oct,
  journal = {Magnetic Resonance Imaging},
  volume = {82},
  pages = {31--41},
  doi = {10.1016/j.mri.2021.06.017},
  lccn = {4}
}

@inproceedings{Ju2013,
  title = {Image Segmentation Based on Adaptive Threshold Edge Detection and Mean Shift},
  booktitle = {2013 IEEE 4th International Conference on Software Engineering and Service Science},
  author = {Ju, Zengwei and Zhou, Jingli and Wang, Xian and Shu, Qin},
  year = {2013},
  month = may,
  pages = {385--388},
  doi = {10.1109/ICSESS.2013.6615330}
}

@inproceedings{Kruger2024,
  title = {Deep Learning-Based Pulmonary Artery Surface Mesh Generation},
  booktitle = {Statistical Atlases and Computational Models of the Heart. Regular and CMRxRecon Challenge Papers},
  author = {Kr{\"u}ger, Nina and Br{\"u}ning, Jan and Goubergrits, Leonid and Ivantsits, Matthias and Walczak, Lars and Falk, Volkmar and Dreger, Henryk and K{\"u}hne, Titus and Hennemuth, Anja},
  editor = {Camara, Oscar and {Puyol-Ant{\'o}n}, Esther and Sermesant, Maxime and Suinesiaputra, Avan and Tao, Qian and Wang, Chengyan and Young, Alistair},
  year = {2024},
  pages = {140--151},
  publisher = {Springer Nature Switzerland},
  address = {Cham},
  doi = {10.1007/978-3-031-52448-6_14},
  isbn = {978-3-031-52448-6}
}

@article{Leyendecker2009,
  title = {Magnetic Resonance Urography},
  author = {Leyendecker, John R and Gianini, John W},
  year = {2009},
  month = jul,
  journal = {Abdominal imaging},
  volume = {34},
  number = {4},
  pages = {527--540},
  doi = {10.1007/s00261-008-9403-9},
  pmid = {18463916}
}

@incollection{Lorensen1998a,
  title = {Marching Cubes: A High Resolution 3D Surface Construction Algorithm},
  shorttitle = {Marching Cubes},
  booktitle = {Seminal Graphics},
  author = {Lorensen, William E. and Cline, Harvey E.},
  year = {1998},
  month = jul,
  pages = {347--353},
  publisher = {ACM},
  address = {New York, NY, USA},
  doi = {10.1145/280811.281026},
  isbn = {978-1-58113-052-2}
}

@inproceedings{Ma2021,
  title = {PialNN: A Fast Deep Learning Framework for Cortical Pial Surface Reconstruction},
  shorttitle = {PialNN},
  booktitle = {Machine Learning in Clinical Neuroimaging: 4th International Workshop, MLCN 2021, Held in Conjunction with MICCAI 2021, Strasbourg, France, September 27, 2021, Proceedings},
  author = {Ma, Qiang and Robinson, Emma C. and Kainz, Bernhard and Rueckert, Daniel and Alansary, Amir},
  year = {2021},
  month = sep,
  pages = {73--81},
  publisher = {Springer-Verlag},
  address = {Berlin, Heidelberg},
  doi = {10.1007/978-3-030-87586-2_8},
  isbn = {978-3-030-87585-5}
}

@article{Manjula2017,
  title = {Image Edge Detection and Segmentation by Using Histogram Thresholding Method},
  author = {Manjula, {\relax Dr}. V.S.},
  year = {2017},
  month = aug,
  journal = {International Journal of Engineering Research and Applications},
  volume = {07},
  number = {08},
  pages = {10--16},
  doi = {10.9790/9622-0708011016}
}

@inproceedings{Mescheder2019,
  title = {Occupancy Networks: Learning 3D Reconstruction in Function Space},
  shorttitle = {Occupancy Networks},
  booktitle = {Proceedings of the IEEE/CVF Conference on Computer Vision and Pattern Recognition},
  author = {Mescheder, Lars and Oechsle, Michael and Niemeyer, Michael and Nowozin, Sebastian and Geiger, Andreas},
  year = {2019},
  pages = {4460--4470}
}

@inproceedings{Moench2010,
  title = {Staircase-Aware Smoothing of Medical Surface Meshes},
  booktitle = {Proceedings of the 2nd Eurographics Conference on Visual Computing for Biology and Medicine},
  author = {Moench, T. and Adler, S. and Preim, B.},
  year = {2010},
  month = jul,
  series = {EG VCBM'10},
  pages = {83--90},
  publisher = {Eurographics Association},
  address = {Goslar, DEU},
  isbn = {978-3-905674-28-6}
}

@article{Morais2024,
  title = {Kidney Collecting System Anatomy Applied to Endourology - a Narrative Review},
  author = {Morais, Ana Raquel M. and Favorito, Luciano A. and Sampaio, Francisco J. B.},
  year = {2024},
  journal = {International Braz J Urol: Official Journal of the Brazilian Society of Urology},
  volume = {50},
  number = {2},
  pages = {164--177},
  doi = {10.1590/S1677-5538.IBJU.2024.9901},
  pmcid = {PMC10953598},
  pmid = {38386787}
}

@article{Nielson2003,
  title = {On Marching Cubes},
  author = {Nielson, G.M.},
  year = {2003},
  month = jul,
  journal = {IEEE Transactions on Visualization and Computer Graphics},
  volume = {9},
  number = {3},
  pages = {283--297},
  doi = {10.1109/TVCG.2003.1207437},
  lccn = {1}
}

@inproceedings{Pan2019,
  title = {Deep Mesh Reconstruction From Single RGB Images via Topology Modification Networks},
  booktitle = {Proceedings of the IEEE/CVF International Conference on Computer Vision},
  author = {Pan, Junyi and Han, Xiaoguang and Chen, Weikai and Tang, Jiapeng and Jia, Kui},
  year = {2019},
  pages = {9964--9973}
}

@inproceedings{Park2019a,
  title = {DeepSDF: Learning Continuous Signed Distance Functions for Shape Representation},
  shorttitle = {DeepSDF},
  booktitle = {2019 IEEE/CVF Conference on Computer Vision and Pattern Recognition (CVPR)},
  author = {Park, Jeong Joon and Florence, Peter and Straub, Julian and Newcombe, Richard and Lovegrove, Steven},
  year = {2019},
  month = jun,
  pages = {165--174},
  publisher = {IEEE},
  address = {Long Beach, CA, USA},
  doi = {10.1109/CVPR.2019.00025},
  copyright = {https://doi.org/10.15223/policy-029},
  isbn = {978-1-72813-293-8}
}

@article{Paszke2019,
  title={Pytorch: An imperative style, high-performance deep learning library},
  author={Paszke, Adam and Gross, Sam and Massa, Francisco and Lerer, Adam and Bradbury, James and Chanan, Gregory and Killeen, Trevor and Lin, Zeming and Gimelshein, Natalia and Antiga, Luca and others},
  journal={Advances in neural information processing systems},
  volume={32},
  year={2019}
}

@article{Qin2021,
  title = {The Influence of Hydronephrosis with Different Degrees on Urodynamics of Renal Pelvis: A Model Study},
  shorttitle = {The Influence of Hydronephrosis with Different Degrees on Urodynamics of Renal Pelvis},
  author = {Qin, Qiqi and Liu, Fan and Duan, Yayu and Han, Pengfei and Zhang, Xuhui},
  year = {2021},
  journal = {Journal of Medical Biomechanics},
  pages = {E877-E882}
}

@misc{Ravi2020,
  title = {Accelerating 3D Deep Learning with PyTorch3D},
  author = {Ravi, Nikhila and Reizenstein, Jeremy and Novotny, David and Gordon, Taylor and Lo, Wan-Yen and Johnson, Justin and Gkioxari, Georgia},
  year = {2020},
  month = jul,
  number = {arXiv:2007.08501},
  eprint = {2007.08501},
  primaryclass = {cs},
  publisher = {arXiv},
  doi = {10.48550/arXiv.2007.08501},
  archiveprefix = {arXiv}
}

@inproceedings{Ronneberger2015,
  title = {U-Net: Convolutional Networks for Biomedical Image Segmentation},
  shorttitle = {U-Net},
  booktitle = {Medical Image Computing and Computer-Assisted Intervention -- MICCAI 2015},
  author = {Ronneberger, Olaf and Fischer, Philipp and Brox, Thomas},
  editor = {Navab, Nassir and Hornegger, Joachim and Wells, William M. and Frangi, Alejandro F.},
  year = {2015},
  pages = {234--241},
  publisher = {Springer International Publishing},
  address = {Cham},
  doi = {10.1007/978-3-319-24574-4_28},
  isbn = {978-3-319-24574-4}
}

@article{Roth2018,
  title = {An Application of Cascaded 3D Fully Convolutional Networks for Medical Image Segmentation},
  author = {Roth, Holger R. and Oda, Hirohisa and Zhou, Xiangrong and Shimizu, Natsuki and Yang, Ying and Hayashi, Yuichiro and Oda, Masahiro and Fujiwara, Michitaka and Misawa, Kazunari and Mori, Kensaku},
  year = {2018},
  month = jun,
  journal = {Computerized Medical Imaging and Graphics},
  volume = {66},
  pages = {90--99},
  doi = {10.1016/j.compmedimag.2018.03.001},
  lccn = {2}
}

@article{Shen2023,
  title = {Flexible Isosurface Extraction for Gradient-Based Mesh Optimization},
  author = {Shen, Tianchang and Munkberg, Jacob and Hasselgren, Jon and Yin, Kangxue and Wang, Zian and Chen, Wenzheng and Gojcic, Zan and Fidler, Sanja and Sharp, Nicholas and Gao, Jun},
  year = {2023},
  month = aug,
  journal = {ACM Transactions on Graphics},
  volume = {42},
  number = {4},
  pages = {1--16},
  doi = {10.1145/3592430},
  lccn = {1}
}

@article{Siddique2021,
  title = {U-Net and Its Variants for Medical Image Segmentation: A Review of Theory and Applications},
  shorttitle = {U-Net and Its Variants for Medical Image Segmentation},
  author = {Siddique, Nahian and Paheding, Sidike and Elkin, Colin P. and Devabhaktuni, Vijay},
  year = {2021},
  journal = {IEEE Access},
  volume = {9},
  pages = {82031--82057},
  doi = {10.1109/ACCESS.2021.3086020},
  copyright = {https://creativecommons.org/licenses/by/4.0/legalcode},
  lccn = {3}
}

@inproceedings{Wang2018a,
  title = {Pixel2Mesh: Generating 3D Mesh Models from Single RGB Images},
  shorttitle = {Pixel2Mesh},
  booktitle = {Proceedings of the European Conference on Computer Vision (ECCV)},
  author = {Wang, Nanyang and Zhang, Yinda and Li, Zhuwen and Fu, Yanwei and Liu, Wei and Jiang, Yu-Gang},
  year = {2018},
  pages = {52--67}
}

@inproceedings{Wickramasinghe2020,
  title = {Voxel2Mesh: 3D Mesh Model Generation from Volumetric Data},
  shorttitle = {Voxel2Mesh},
  booktitle = {Medical Image Computing and Computer Assisted Intervention -- MICCAI 2020: 23rd International Conference, Lima, Peru, October 4--8, 2020, Proceedings, Part IV},
  author = {Wickramasinghe, Udaranga and Remelli, Edoardo and Knott, Graham and Fua, Pascal},
  year = {2020},
  month = oct,
  pages = {299--308},
  publisher = {Springer-Verlag},
  address = {Berlin, Heidelberg},
  doi = {10.1007/978-3-030-59719-1_30},
  isbn = {978-3-030-59718-4}
}

@article{Xu2019,
  title={Disn: Deep implicit surface network for high-quality single-view 3d reconstruction},
  author={Xu, Qiangeng and Wang, Weiyue and Ceylan, Duygu and Mech, Radomir and Neumann, Ulrich},
  journal={Advances in neural information processing systems},
  volume={32},
  year={2019}
}

@inproceedings{Zhou2018,
  title = {UNet++: A Nested U-Net Architecture for Medical Image Segmentation},
  shorttitle = {UNet++},
  booktitle = {Deep Learning in Medical Image Analysis and Multimodal Learning for Clinical Decision Support},
  author = {Zhou, Zongwei and Rahman Siddiquee, Md Mahfuzur and Tajbakhsh, Nima and Liang, Jianming},
  editor = {Stoyanov, Danail and Taylor, Zeike and Carneiro, Gustavo and {Syeda-Mahmood}, Tanveer and Martel, Anne and {Maier-Hein}, Lena and Tavares, Jo{\~a}o Manuel R.S. and Bradley, Andrew and Papa, Jo{\~a}o Paulo and Belagiannis, Vasileios and Nascimento, Jacinto C. and Lu, Zhi and Conjeti, Sailesh and Moradi, Mehdi and Greenspan, Hayit and Madabhushi, Anant},
  year = {2018},
  pages = {3--11},
  publisher = {Springer International Publishing},
  address = {Cham},
  doi = {10.1007/978-3-030-00889-5_1},
  isbn = {978-3-030-00889-5}
}

@article{jang2022numerical,
  title={Numerical investigation of urethra flow characteristics in benign prostatic hyperplasia},
  author={Jang, Kyeong Sik and Kim, Jin Wook and Ryu, Jaiyoung},
  journal={Computer Methods and Programs in Biomedicine},
  volume={224},
  pages={106978},
  year={2022},
  publisher={Elsevier}
}

@article{inman2013impact,
  title={The impact of temperature and urinary constituents on urine viscosity and its relevance to bladder hyperthermia treatment},
  author={Inman, Brant A and Etienne, Wiguins and Rubin, Rainier and Owusu, Richmond A and Oliveira, Tiago R and Rodriques, Dario B and Maccarini, Paolo F and Stauffer, Paul R and Mashal, Alireza and Dewhirst, Mark W},
  journal={International Journal of Hyperthermia},
  volume={29},
  number={3},
  pages={206--210},
  year={2013},
  publisher={Taylor \& Francis}
}

@article{maralescu2022non,
  title={Non-invasive evaluation of kidney elasticity and viscosity in a healthy cohort},
  author={Maralescu, Felix-Mihai and Bende, Felix and Sporea, Ioan and Popescu, Alina and Sirli, Roxana and Schiller, Adalbert and Petrica, Ligia and Miutescu, Bogdan and Borlea, Andreea and Popa, Alexandru and others},
  journal={Biomedicines},
  volume={10},
  number={11},
  pages={2859},
  year={2022},
  publisher={MDPI}
}

@article{antiga2002geometric,
  title={Geometric reconstruction for computational mesh generation of arterial bifurcations from CT angiography},
  author={Antiga, Luca and Ene-Iordache, Bogdan and Caverni, Lionello and Cornalba, Gian Paolo and Remuzzi, Andrea},
  journal={Computerized Medical Imaging and Graphics},
  volume={26},
  number={4},
  pages={227--235},
  year={2002},
  publisher={Elsevier}
}

@article{leinster2008euler,
  title={The Euler characteristic of a category},
  author={Leinster, Tom},
  journal={Documenta Mathematica},
  volume={13},
  pages={21--49},
  year={2008}
}

@article{charton2021mesh,
  title={Mesh repairing using topology graphs},
  author={Charton, Jerome and Baek, Stephen and Kim, Youngjun},
  journal={Journal of Computational Design and Engineering},
  volume={8},
  number={1},
  pages={251--267},
  year={2021},
  publisher={Oxford University Press}
}

@article{bacciaglia2021surface,
  title={Surface smoothing for topological optimized 3D models},
  author={Bacciaglia, Antonio and Ceruti, Alessandro and Liverani, Alfredo},
  journal={Structural and Multidisciplinary Optimization},
  volume={64},
  number={6},
  pages={3453--3472},
  year={2021},
  publisher={Springer}
}

@article{pak2023patient,
  title={Patient-specific heart geometry modeling for solid biomechanics using deep learning},
  author={Pak, Daniel H and Liu, Minliang and Kim, Theodore and Liang, Liang and Caballero, Andres and Onofrey, John and Ahn, Shawn S and Xu, Yilin and McKay, Raymond and Sun, Wei and others},
  journal={IEEE transactions on medical imaging},
  year={2023},
  publisher={IEEE}
}

@article{ferdian2023cerebrovascular,
  title={Cerebrovascular super-resolution 4D flow MRI--sequential combination of resolution enhancement by deep learning and physics-informed image processing to non-invasively quantify intracranial velocity, flow, and relative pressure},
  author={Ferdian, Edward and Marlevi, David and Schollenberger, Jonas and Aristova, Maria and Edelman, Elazer R and Schnell, Susanne and Figueroa, C Alberto and Nordsletten, DA and Young, Alistair A},
  journal={Medical Image Analysis},
  volume={88},
  pages={102831},
  year={2023},
  publisher={Elsevier}
}

@article{decroocq2023modeling,
  title={Modeling and hexahedral meshing of cerebral arterial networks from centerlines},
  author={Decroocq, M{\'e}ghane and Frindel, Carole and Roug{\'e}, Pierre and Ohta, Makoto and Lavou{\'e}, Guillaume},
  journal={Medical image analysis},
  volume={89},
  pages={102912},
  year={2023},
  publisher={Elsevier}
}

@inproceedings{aase2023graph,
  title={Graph Convolutional Neural Networks for Automated Echocardiography View Recognition: A Holistic Approach},
  author={Aase, B},
  booktitle={Simplifying Medical Ultrasound: 4th International Workshop, ASMUS 2023, Held in Conjunction with MICCAI 2023, Vancouver, BC, Canada, October 8, 2023, Proceedings},
  volume={14337},
  pages={44},
  year={2023},
  organization={Springer Nature}
}

@article{yotter2011topological,
  title={Topological correction of brain surface meshes using spherical harmonics},
  author={Yotter, Rachel Aine and Dahnke, Robert and Thompson, Paul M and Gaser, Christian},
  journal={Human brain mapping},
  volume={32},
  number={7},
  pages={1109--1124},
  year={2011},
  publisher={Wiley Online Library}
}

@article{wyburd2024anatomically,
  title={Anatomically plausible segmentations: Explicitly preserving topology through prior deformations},
  author={Wyburd, Madeleine K and Dinsdale, Nicola K and Jenkinson, Mark and Namburete, Ana IL},
  journal={Medical Image Analysis},
  pages={103222},
  year={2024},
  publisher={Elsevier}
}

@article{Lebrat2021,
  title={CorticalFlow: A diffeomorphic mesh transformer network for cortical surface reconstruction},
  author={Lebrat, Leo and Santa Cruz, Rodrigo and De Gournay, Frederic and Fu, Darren and Bourgeat, Pierrick and Fripp, Jurgen and Fookes, Clinton and Salvado, Olivier},
  journal={Advances in Neural Information Processing Systems},
  volume={34},
  pages={29491--29505},
  year={2021}
}

@inproceedings{CorticalFlowPP2022,
  title={CorticalFlow++: Boosting cortical surface reconstruction accuracy, regularity, and interoperability},
  author={Santa Cruz, Rodrigo and Lebrat, L{\'e}o and Fu, Darren and Bourgeat, Pierrick and Fripp, Jurgen and Fookes, Clinton and Salvado, Olivier},
  booktitle={International Conference on Medical Image Computing and Computer-Assisted Intervention},
  pages={496--505},
  year={2022},
  organization={Springer}
}

@article{Ma2022CortexODE,
  title={CortexODE: Learning cortical surface reconstruction by neural ODEs},
  author={Ma, Qiang and Li, Liu and Robinson, Emma C and Kainz, Bernhard and Rueckert, Daniel and Alansary, Amir},
  journal={IEEE Transactions on Medical Imaging},
  volume={42},
  number={2},
  pages={430--443},
  year={2022},
  publisher={IEEE}
}

@article{Kong2021WholeHeart,
  title={A deep-learning approach for direct whole-heart mesh reconstruction},
  author={Kong, Fanwei and Wilson, Nathan and Shadden, Shawn},
  journal={Medical image analysis},
  volume={74},
  pages={102222},
  year={2021},
  publisher={Elsevier}
}

@inproceedings{pak2021distortion,
  title={Distortion energy for deep learning-based volumetric finite element mesh generation for aortic valves},
  author={Pak, Daniel H and Liu, Minliang and Kim, Theodore and Liang, Liang and McKay, Raymond and Sun, Wei and Duncan, James S},
  booktitle={International Conference on Medical Image Computing and Computer-Assisted Intervention},
  pages={485--494},
  year={2021},
  organization={Springer}
}

@article{narayanan2024linflo,
  title={LinFlo-Net: A two-stage deep learning method to generate simulation ready meshes of the heart},
  author={Narayanan, Arjun and Kong, Fanwei and Shadden, Shawn},
  journal={Journal of Biomechanical Engineering},
  volume={146},
  number={7},
  pages={071005},
  year={2024},
  publisher={American Society of Mechanical Engineers}
}

@inproceedings{lee2015deeply,
  title={Deeply-supervised nets},
  author={Lee, Chen-Yu and Xie, Saining and Gallagher, Patrick and Zhang, Zhengyou and Tu, Zhuowen},
  booktitle={Artificial intelligence and statistics},
  pages={562--570},
  year={2015},
  organization={Pmlr}
}

@article{wang2015training,
  title={Training deeper convolutional networks with deep supervision},
  author={Wang, Liwei and Lee, Chen-Yu and Tu, Zhuowen and Lazebnik, Svetlana},
  journal={arXiv preprint arXiv:1505.02496},
  year={2015}
}

@inproceedings{zhang2018deep,
  title={Deep supervision with additional labels for retinal vessel segmentation task},
  author={Zhang, Yishuo and Chung, Albert CS},
  booktitle={International conference on medical image computing and computer-assisted intervention},
  pages={83--91},
  year={2018},
  organization={Springer}
}

@article{attene2010lightweight,
  title={A lightweight approach to repairing digitized polygon meshes},
  author={Attene, Marco},
  journal={The visual computer},
  volume={26},
  number={11},
  pages={1393--1406},
  year={2010},
  publisher={Springer}
}

@book{botsch2010polygon,
  title={Polygon mesh processing},
  author={Botsch, Mario and Kobbelt, Leif and Pauly, Mark and Alliez, Pierre and L{\'e}vy, Bruno},
  year={2010},
  publisher={CRC press}
}

@inproceedings{vetsch2022neuralmeshing,
  title={Neuralmeshing: Differentiable meshing of implicit neural representations},
  author={Vetsch, Mathias and Lombardi, Sandro and Pollefeys, Marc and Oswald, Martin R},
  booktitle={DAGM German Conference on Pattern Recognition},
  pages={317--333},
  year={2022},
  organization={Springer}
}

@article{soboleva2023prs,
  title={PRS: Sharp Feature Priors for Resolution-Free Surface Remeshing},
  author={Soboleva, Natalia and Gorbunova, Olga and Ivanova, Maria and Burnaev, Evgeny and Nie{\ss}ner, Matthias and Zorin, Denis and Artemov, Alexey},
  journal={arXiv preprint arXiv:2311.18494},
  year={2023}
}

@article{fu2024lfs,
  title={LFS-aware surface reconstruction from unoriented 3D point clouds},
  author={Fu, Rao and Hormann, Kai and Alliez, Pierre},
  journal={IEEE Transactions on Multimedia},
  year={2024},
  publisher={IEEE}
}

@article{kong2022learning,
  title={Learning whole heart mesh generation from patient images for computational simulations},
  author={Kong, Fanwei and Shadden, Shawn C},
  journal={IEEE Transactions on Medical Imaging},
  volume={42},
  number={2},
  pages={533--545},
  year={2022},
  publisher={IEEE}
}

@article{isensee2018nnu,
  title={nnu-net: Self-adapting framework for u-net-based medical image segmentation},
  author={Isensee, Fabian and Petersen, Jens and Klein, Andre and Zimmerer, David and Jaeger, Paul F and Kohl, Simon and Wasserthal, Jakob and Koehler, Gregor and Norajitra, Tobias and Wirkert, Sebastian and others},
  journal={arXiv preprint arXiv:1809.10486},
  year={2018}
}

@inproceedings{milletari2016v,
  title={V-net: Fully convolutional neural networks for volumetric medical image segmentation},
  author={Milletari, Fausto and Navab, Nassir and Ahmadi, Seyed-Ahmad},
  booktitle={2016 fourth international conference on 3D vision (3DV)},
  pages={565--571},
  year={2016},
  organization={Ieee}
}

@article{hernandez2025sr,
  title={SR-CurvANN: Advancing 3D surface reconstruction through curvature-aware neural networks},
  author={Hern{\'a}ndez-Bautista, Marina and Melero, Francisco J},
  journal={Computers \& Graphics},
  pages={104260},
  year={2025},
  publisher={Elsevier}
}

@inproceedings{zhang2025high,
  title={High-Fidelity Lightweight Mesh Reconstruction from Point Clouds},
  author={Zhang, Chen and Wang, Wentao and Li, Ximeng and Liao, Xinyao and Su, Wanjuan and Tao, Wenbing},
  booktitle={Proceedings of the Computer Vision and Pattern Recognition Conference},
  pages={11739--11748},
  year={2025}
}

@article{yang2025non,
  title={Non-invasive urine flow dynamics characterization of pediatric hydronephrosis based on deep learning and computational fluid dynamics},
  author={Yang, Mingjing and Zhu, Zhanpeng and Huang, Liqin and Sun, Haoran and Lin, Xingtao and Li, Nuoxi and Pan, Lin and Lin, Shan and Ding, Wangbin},
  journal={Computer Methods and Programs in Biomedicine},
  pages={109077},
  year={2025},
  publisher={Elsevier}
}

\end{document}